\RequirePackage{fix-cm}
\documentclass[twocolumn]{svjour3}          % twocolumn
\smartqed  % flush right qed marks, e.g. at end of proof

\usepackage{times}
\usepackage{epsfig}
\usepackage{graphicx}
\usepackage[fleqn]{amsmath}
\usepackage{amssymb}
\usepackage{booktabs}
\usepackage{multirow}
\usepackage{url}
\usepackage[numbers]{natbib}
\usepackage{color}
\usepackage{float}

\begin{document}

\title{Semantic Understanding of Scenes through the ADE20K Dataset}



\author{Bolei Zhou \and
       Hang Zhao \and
       Xavier Puig \and
       Tete Xiao \and
       Sanja Fidler \and
       Adela Barriuso \and
       Antonio Torralba
}


\institute{
B. Zhou \at
Department of Information Engineering, the Chinese University of Hong Kong, Hong Kong.
\and
H. Zhao, X. Puig, A. Barriuso,  A. Torralba \at
              Computer Science and Artificial Intelligence Laboratory, Massachusetts Institute of Technology, USA.
              % Department of Electrical Engineering and Computer Science, Massachusetts Institute of Technology, USA.  \\
           \and
           T. Xiao \at School of Electronic Engineering and Computer Science, Peking University, China.
           \and
           S. Fidler \at
             Department of Computer Science, University of Toronto, Canada.
}


\maketitle

\begin{abstract}
Semantic understanding of visual scenes is one of the holy grails of computer vision. Despite efforts of the community in data collection, there are still few image datasets covering a wide range of scenes and object categories with pixel-wise annotations for scene understanding. In this work, we present a densely annotated dataset \textit{ADE20K}, which spans diverse annotations of scenes, objects, parts of objects, and in some cases even parts of parts. Totally there are 25k images of the complex everyday scenes containing a variety of objects in their natural spatial context. On average there are 19.5 instances and 10.5 object classes per image. Based on ADE20K, we construct benchmarks for scene parsing and instance segmentation. We provide baseline performances on both of the benchmarks and re-implement the state-of-the-art models for open source. We further evaluate the effect of synchronized batch normalization and find that a reasonably large batch size is crucial for the semantic segmentation performance. We show that the networks trained on ADE20K are able to segment a wide variety of scenes and objects\footnote{Dataset is available at \url{http://groups.csail.mit.edu/vision/datasets/ADE20K}. \\Pretrained models and code are released at \url{https://github.com/CSAILVision/semantic-segmentation-pytorch}}.

%Scene parsing, or recognizing and segmenting objects and stuff in an image, is one of the key problems in computer vision.  Despite the community's efforts in data collection, there are still few image datasets covering a wide range of scenes and object categories with dense and detailed annotations for scene parsing. In this paper, we introduce and analyze the ADE20K dataset, spanning diverse annotations of scenes, objects, parts of objects, and in some cases even parts of parts. A generic network design called Cascade Segmentation Module is then proposed to enable the segmentation networks to parse a scene into stuff, objects, and object parts in a cascade. We evaluate the proposed module integrated within two existing semantic segmentation networks, yielding significant improvements for scene parsing. We further show that the scene parsing networks trained on ADE20K can be applied to a wide variety of scenes and objects\footnote{Dataset is available at \url{http://groups.csail.mit.edu/vision/datasets/ADE20K/}.}.
\end{abstract}

\begin{figure*}
\centering
\includegraphics[width=1\textwidth]{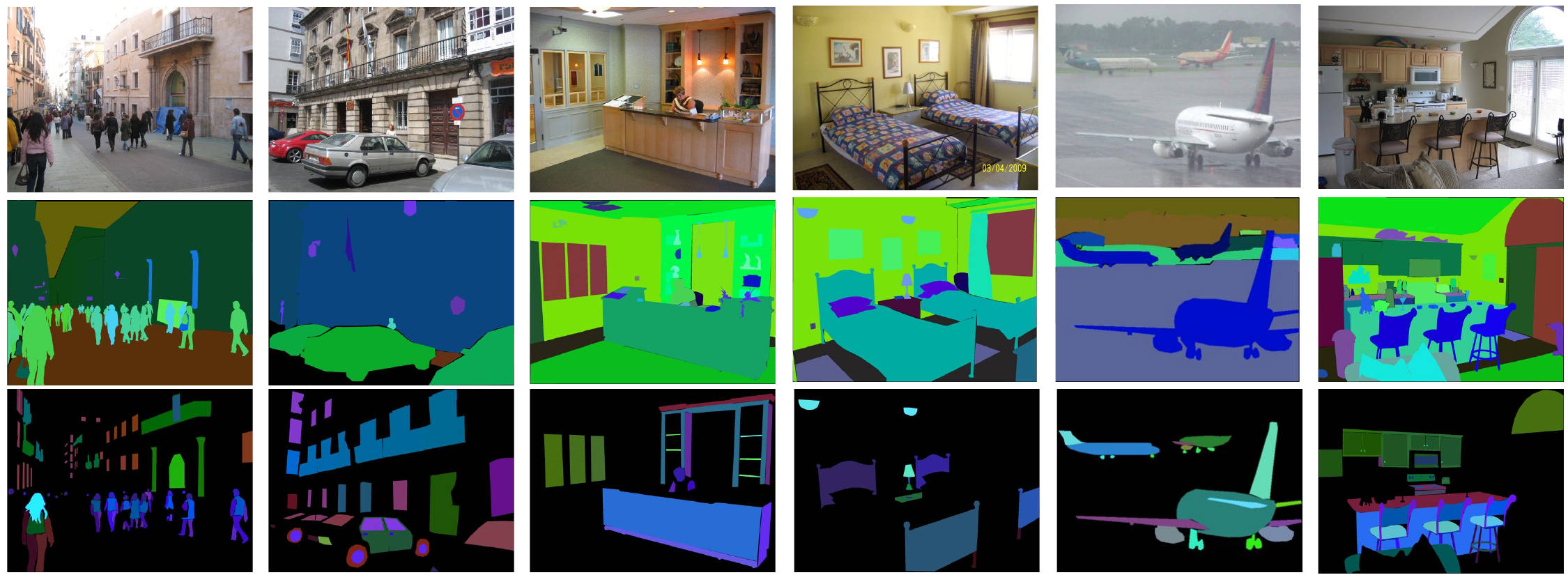}
\caption{Images in ADE20K dataset are densely annotated in detail with objects and parts. The first row shows the sample images, the second row shows the annotation of objects, and the third row shows the annotation of object parts. The color scheme both encodes the object categories and object instances, that different object categories have large color difference while different instances from the same object category have small color difference (e.g., different person instances in first image have slightly different colors).}
%ADE20K dataset that has densely and detailed annotated images with objects and parts. The first row includes the sample images, the second row includes the annotation of objects, and the third row includes the annotation of object parts.}
\label{fig:teaser}
\end{figure*}

\section{Introduction}

Semantic understanding of visual scenes is one of the holy grails of computer vision. The emergence of large-scale image datasets like ImageNet \cite{ILSVRC15}, COCO \cite{lin2014microsoft} and Places \cite{zhou2014learning}, along with the rapid development of the deep convolutional neural network (CNN) approaches, have brought great advancements to visual scene understanding. Nowadays, given a visual scene of a living room, a robot equipped with a trained CNN can accurately predict the scene category. However, to freely navigate in the scene and manipulate the objects inside, the robot has far more information to extract from the input image: It needs to recognize and localize not only the objects like sofa, table, and TV, but also their parts, e.g., a seat of a chair or a handle of a cup, to allow proper manipulation, as well as to segment the stuff like floor, wall and ceiling for spatial navigation.

Recognizing and segmenting objects and stuff at pixel level remains one of the key problems in scene understanding. Going beyond the image-level recognition, the pixel-level scene understanding requires a much denser annotation of scenes with a large set of objects. However, the current datasets have a limited number of objects (e.g., COCO~\cite{lin2014microsoft}, Pascal~\cite{everingham2010pascal}) and in many cases those objects are not the most common objects one encounters in the world (like frisbees or baseball bats), or the datasets only cover a limited set of scenes (e.g., Cityscapes~\cite{Cordts2015Cvprw}). Some notable exceptions are Pascal-Context~\cite{mottaghi2014role} and the SUN database~\cite{xiao2010sun}. However, Pascal-Context still contains scenes primarily focused on 20 object classes, while SUN has noisy labels at the object level. 

%Devising real-world recognition applications, such as a robot navigating the world or an online image search engine, will require recognizing a very large set of common objects. 
The motivation of this work is to collect a dataset that has densely annotated images (every pixel has a semantic label) with a large and an unrestricted open vocabulary. The images in our dataset are manually segmented in great detail, covering a diverse set of scenes, object and object part categories. The challenge for collecting such annotations is finding reliable annotators, as well as the fact that labeling is difficult if the class list is not defined in advance. On the other hand, open vocabulary naming also suffers from naming inconsistencies across different annotators. In contrast, our dataset was annotated by a single expert annotator, providing extremely detailed and exhaustive image annotations. On average, our annotator labeled 29 annotation segments per image, compared to the 16 segments per image labeled by external annotators (like workers from Amazon Mechanical Turk). Furthermore, the data consistency and quality are much higher than that of external annotators. Fig.~\ref{fig:teaser} shows examples from our dataset. 

The preliminary result of this work is published at \cite{zhou2017scene}. Compared to the previous conference paper, we include more description of the dataset, more baseline results on the scene parsing benchmark, the introduction of the new instance segmentation benchmark and its baseline results, as well as the effect of synchronized batch norm and the joint training of objects and parts. We also include the contents of the Places Challenges we hosted at ECCV'16 and ICCV'17 and the analysis on the challenge results.

%To establish a quality estimate of our dataset, we ``benchmarked'' our annotator against two (trained) AMT-like external annotators. The external annotators labeled on average 16 segments per image while our annotator provided 29 segments per image. Furthermore, the consistency among the external annotators was much lower than that of our annotator.  

The sections of this work are organized as follows. In Sec.\ref{section:dataset_construction} we describe the construction of the ADE20K dataset and its statistics. In Sec.\ref{section:benchmarks} we introduce the two pixel-wise scene understanding benchmarks we build upon ADE20K: scene parsing and instance segmentation. We train and evaluate several baseline networks on the benchmarks. We also re-implement and open-source several state-of-the-art scene parsing models and evaluate the effect of batch normalization size. In Sec.\ref{section:challenges} we introduce the Places Challenges at ECCV'16 and ICCV'17 based on the benchmarks of the ADE20K, as well as the qualitative and quantitative analysis on the challenge results. In Sec.\ref{section:jointtraining} we train network jointly to segment objects and their parts. Sec.\ref{section:applications} explores the applications of the scene parsing networks to the hierarchical semantic segmentation and automatic scene content removal. Sec.\ref{section:conclusion} concludes this work.

\subsection{Related work}

Many datasets have been collected for the purpose of semantic understanding of scenes. We review the datasets according to the level of details of their annotations, then briefly go through the previous work of semantic segmentation networks.

{\bf Object classification/detection datasets.} Most of the large-scale datasets typically only contain labels at the image level or provide bounding boxes. Examples include ImageNet~\citep{ILSVRC15}, Pascal~\citep{everingham2010pascal}, and KITTI~\citep{Geiger2012CVPR}. ImageNet has the largest set of classes, but contains relatively simple scenes. Pascal and KITTI are more challenging and have more objects per image, however, their classes and scenes are more constrained. 

{\bf Semantic segmentation datasets.} Existing datasets with pixel-level labels typically provide annotations only for a subset of foreground objects (20 in PASCAL VOC~\citep{everingham2010pascal} and 91 in Microsoft COCO~\citep{lin2014microsoft}). Collecting dense annotations where all pixels are labeled is much more challenging. Such efforts include Pascal-Context~\citep{mottaghi2014role}, NYU Depth V2~\citep{Silberman:ECCV12}, SUN database~\citep{xiao2010sun}, SUN RGB-D dataset~\citep{song2015sun}, CityScapes dataset~\citep{Cordts2015Cvprw}, and OpenSurfaces~\citep{bell13opensurfaces,bell2015material}. Recently COCO stuff dataset \cite{cocostuff} provides additional stuff segmentation complementary to the 80 object categories in COCO dataset, while COCO attributes dataset \cite{cocoattribute} annotates attributes for some objects in COCO dataset. Such a dataset with progressive enhancement of diverse annotations over the years makes great progress to the modern development of image dataset.

{\bf Datasets with objects, parts and attributes.} Two datasets were released that go beyond the typical labeling setup by also providing pixel-level annotation for the object parts, i.e., Pascal-Part dataset~\citep{chen2014parts}, or material classes, i.e., OpenSurfaces~\citep{bell13opensurfaces,bell2015material}. We advance this effort by collecting very high-resolution imagery of a much wider selection of scenes, containing a large set of object classes per image. We annotated both stuff and object classes, for which we additionally annotated their parts, and parts of these parts. We believe that our dataset, ADE20K, is one of the most comprehensive datasets of its kind. We provide a comparison between datasets in Sec.~\ref{sec:datasetcomparison}.

{\bf Semantic segmentation models.} With the success of convolutional neural networks (CNN) for image classification \citep{krizhevsky2012imagenet}, there is growing interest for semantic pixel-wise labeling using CNNs with dense output, such as the fully CNN \citep{long2015fully}, deconvolutional neural networks \citep{noh2015learning}, encoder-decoder SegNet \citep{badrinarayanan2015segnet}, multi-task network cascades \citep{dai2015instance}, and DilatedVGG \citep{CP2016Deeplab, YuKoltun2016}. They are benchmarked on Pascal dataset with impressive performance on segmenting the 20 object classes. Some of them \citep{long2015fully,badrinarayanan2015segnet} are evaluated on Pascal Context~\citep{mottaghi2014role} or SUN RGB-D dataset~\citep{song2015sun} to show the capability to segment more object classes in scenes. Joint stuff and object segmentation is explored in~\citep{dai2015convolutional} which uses pre-computed superpixels and feature masking to represent stuff. Cascade of instance segmentation and categorization has been explored in \citep{dai2015instance}. A multiscale pyramid pooling module is proposed to improve the scene parsing \cite{zhao2017pyramid}. A recent multi-task segmentation network UperNet is proposed to segment visual concepts from different levels \cite{xiao2018unified}. 

\section{ADE20K: Fully Annotated Image Dataset}
\label{section:dataset_construction}
In this section, we describe the construction of our ADE20K dataset and analyze its statistics. 

\begin{figure*}
\centering
\includegraphics[width=0.8\textwidth]{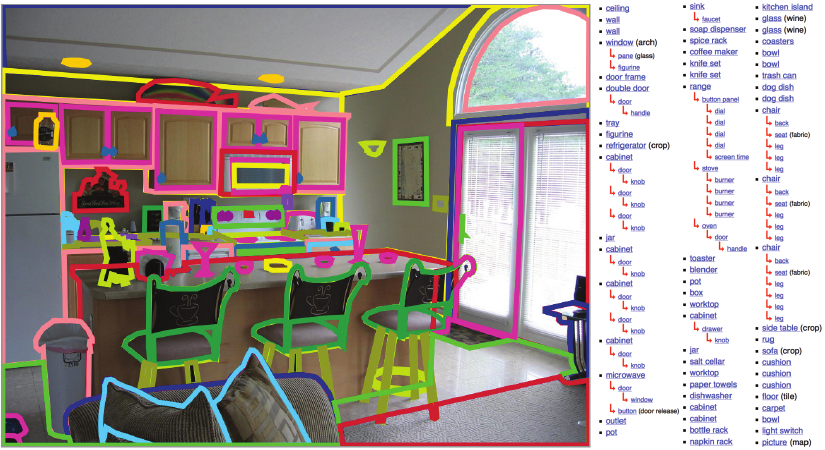}
\caption{Annotation interface, the list of the objects and their associated parts in the image.}
\label{fig:labelme}
\end{figure*}

\begin{figure*}
\centering
\includegraphics[width=0.8\textwidth]{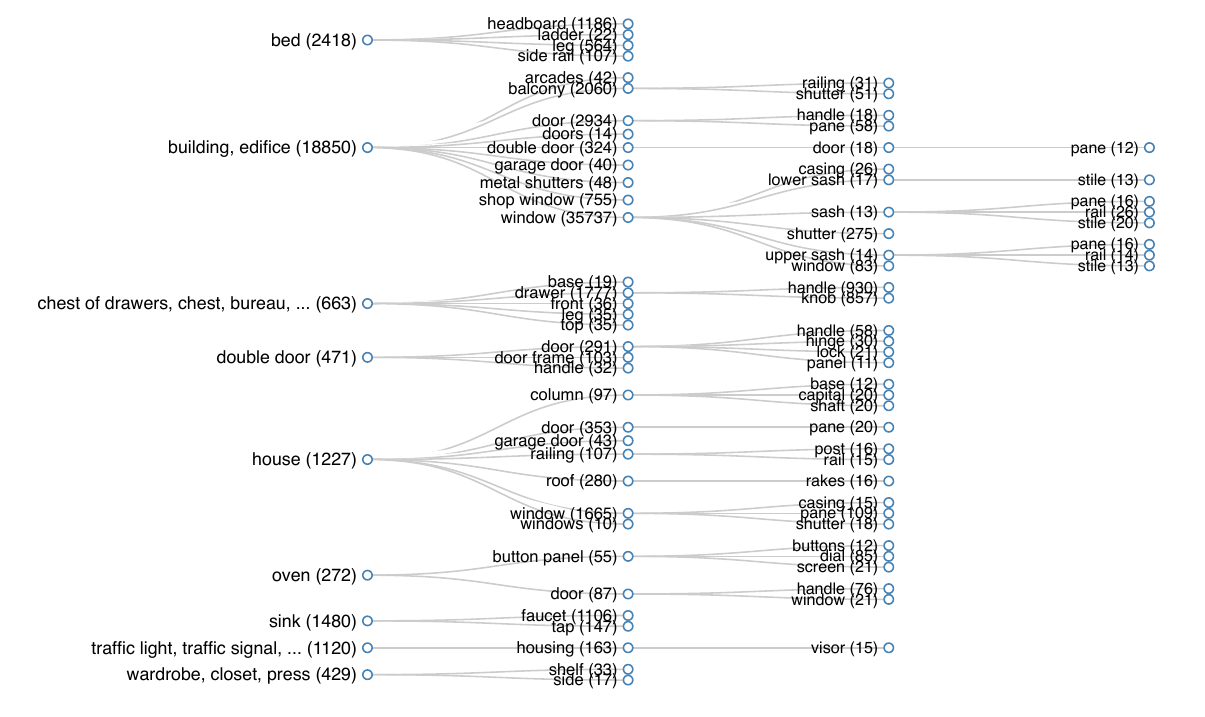}
\caption{Section of the relation tree of objects and parts for the dataset. Each number indicates the number of instances for each object. The full relation tree is available at the dataset webpage.}
\label{fig:tree_fig}
\end{figure*}

%where part annotations are not exhaustive in the training set. %The split into training, validation and test sets is done so as to cover the same distribution over scenes.  

\subsection{Image annotation}

For our dataset, we are interested in having a diverse set of scenes with dense annotations of all the visual concepts present. The visual concepts could be 1) discrete object which is a thing with a well-defined shape, e.g., car, person, 2) stuff which contains amorphous background regions, e.g., grass, sky, or 3) object part, which is a component of some existing object instance which has some functional meaning, such as head or leg. Images come from the LabelMe~\citep{russell2008labelme}, SUN datasets~\citep{xiao2010sun}, and Places~\citep{zhou2014learning} and were selected to cover the 900 scene categories defined in the SUN database. Images were annotated by a single expert worker using the LabelMe interface~\citep{russell2008labelme}.  Fig.~\ref{fig:labelme} shows a snapshot of the annotation interface and one fully segmented image. The worker provided three types of annotations: object segments with names, object parts, and attributes. All object instances are segmented independently so that the dataset could be used to train and evaluate detection or segmentation algorithms.

Given that the objects appearing in the dataset are fully annotated, even in the regions where these are occluded, there are multiple areas where the polygons from different regions overlap. In order to convert the annotated polygons into a segmentation mask, we sort objects in an image by depth layers. Background classes like `sky' or `wall' are set as the farthest layers. The rest of objects' depths are set as follows: when a polygon is fully contained inside another polygon, the object from the inner polygon is given a closer depth layer. When objects only partially overlap, we look at the region of intersection between the two polygons, and set as the closest object the one whose polygon has more points in the region of intersection. Once objects have been sorted, the segmentation mask is constructed by iterating over the objects in decreasing depth, ensuring that object parts never occlude whole objects and no object is occluded by its parts.    

Datasets such as COCO~\citep{lin2014microsoft}, Pascal~\citep{everingham2010pascal} or Cityscape~\citep{Cordts2015Cvprw} start by defining a set of object categories of interest. However, when labeling all the objects in a scene, working with a predefined list of objects is not possible as new categories appear frequently (see fig.~\ref{fig:stats2}.d). Here, the annotator created a dictionary of visual concepts where new classes were added constantly to ensure consistency in object naming. 

Object parts are associated with object instances. Note that parts can have parts too, and we label these associations as well. For example, the `rim' is a part of a `wheel', which in turn is part of a `car'. A `knob' is a part of a `door' that can be part of a `cabinet'. This part hierarchy in Fig.~\ref{fig:tree_fig} has a depth of 3.

\subsection{Dataset summary}
After annotation, there are $20,210$ images in the training set, $2,000$ images in the validation set, and $3,000$ images in the testing set. There are in total $3,169$ class labels annotated, among them $2,693$ are object and stuff classes, $476$ are object part classes. All the images are exhaustively annotated with objects. Many objects are also annotated with their parts. For each object there is additional information about whether it is occluded or cropped, and other attributes. The images in the validation set are exhaustively annotated with parts, while the part annotations are not exhaustive over the images in the training set. Sample images and annotations from the ADE20K dataset are shown in Fig.~\ref{fig:teaser}. 

\subsection{Annotation consistency}
\label{sec:consistency}
Defining a labeling protocol is relatively easy when the labeling task is restricted to a fixed list of object classes, however it becomes challenging when the class list is open-ended. As the goal is to label all the objects within each image, the list of classes grows unbounded. Many object classes appear only a few times across the entire collection of images. However, those rare object classes cannot be ignored as they might be important elements for the interpretation of the scene. Labeling in these conditions becomes difficult because we need to keep a growing list of all the object classes in order to have a consistent naming across the entire dataset. Despite the best effort of the annotator, the process is not free from noise.

\begin{figure*}
\centering
\includegraphics[width=1\textwidth]{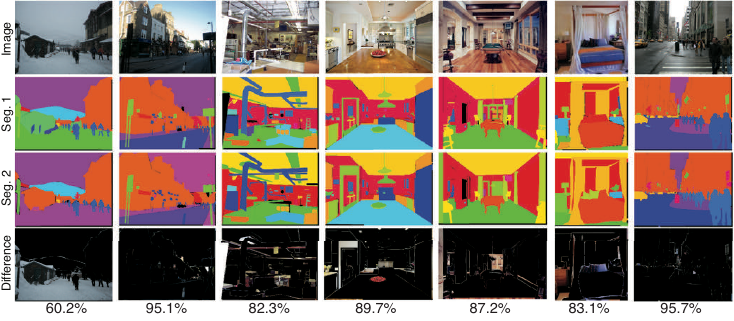}
\caption{Analysis of annotation consistency. Each column shows an image and two segmentations done by the same annotator at different times. Bottom row shows the pixel discrepancy when the two segmentations are subtracted, while the number at the bottom shows the percentage of pixels with the same label. On average across all re-annotated images, $82.4\%$ of pixels got the same label. In the example in the first column the percentage of pixels with the same label is relatively low because the annotator labeled the same region as `snow' and `ground' during the two rounds of annotation. In the third column, there were many objects in the scene and the annotator missed some between the two segmentations.}
\label{fig:consistency}
\end{figure*}

%As shown in Fig.~\ref{fig:consistency}.b, on average, $81.9\%$ of the pixels got the same label (the median is $86\%$). Some of the differences are due to missing objects, naming inconsistency (the image on the left got the label `ground' and `snow'), and differences in the segmentation boundary (e.g., trees can be segmented with different degrees of precision). We further asked two additional annotators to annotate the same set of images. The consistency decreases to $35.6\%$ when comparing with other annotators, mainly due to naming inconsistency.

To analyze the annotation consistency we took a subset of 61 randomly chosen images from the validation set, then asked our annotator to annotate them again (there is a time difference of six months). One expects that there are some differences between the two annotations. A few examples are shown in Fig~\ref{fig:consistency}. On average, $82.4\%$ of the pixels got the same label. The remaining 17.6\% of pixels had some errors for which we grouped into three error types as follows:

\begin{itemize}    
\item   {\bf Segmentation quality}: Variations in the quality of segmentation and outlining of the object boundary. One typical source of error arises when segmenting complex objects such as buildings and trees, which can be segmented with different degrees of precision. This type of error emerges in 5.7\% of the pixels.

\item   {\bf Object naming}: Differences in object naming (due to ambiguity or similarity between concepts, for instance, calling a big car a `car' in one segmentation and a `truck' in the another one, or a `palm tree' a `tree'. This naming issue emerges in 6.0\% of the pixels. These errors can be reduced by defining a very precise terminology, but this becomes much harder with a large growing vocabulary.

\item   {\bf Segmentation quantity}: Missing objects in one of the two segmentations. There is a very large number of objects in each image and some images might be annotated more thoroughly than others. For example, in the third column of Fig.~\ref{fig:consistency} the annotator missed some small objects in different annotations. Missing labels account for 5.9\% of the error pixels. A similar issue existed in segmentation datasets such as the Berkeley Image segmentation dataset \citep{MartinFTM01}.
\end{itemize}

The median error values for the three error types are: 4.8\%, 0.3\% and 2.6\% showing that the mean value is dominated by a few images, and that the most common type of error is segmentation quality. 

To further compare the annotation done by our single expert annotator and the AMT-like annotators, 20 images from the validation set are annotated by two invited external annotators, both with prior experience in image labeling. The first external annotator had 58.5\% of inconsistent pixels compared to the segmentation provided by our annotator, and the second external annotator had 75\% of the inconsistent pixels. Many of these inconsistencies are due to the poor quality of the segmentations provided by external annotators (as it has been observed with AMT which requires multiple verification steps for quality control~\citep{lin2014microsoft}). For the best external annotator (the first one), 7.9\% of pixels have inconsistent segmentations (just slightly worse than our annotator), 14.9\% have inconsistent object naming and 35.8\% of the pixels correspond to missing objects, which is due to the much smaller number of objects annotated by the external annotator in comparison with the ones annotated by our expert annotator. The external annotators labeled on average 16 segments per image while our annotator provided 29 segments per image.

%The external annotators were not shown any example images. So this result shows that the bias is small and that quality can be improved by providing better tools.

\subsection{Dataset statistics}

\begin{figure*}
\centering
\includegraphics[width=1\textwidth]{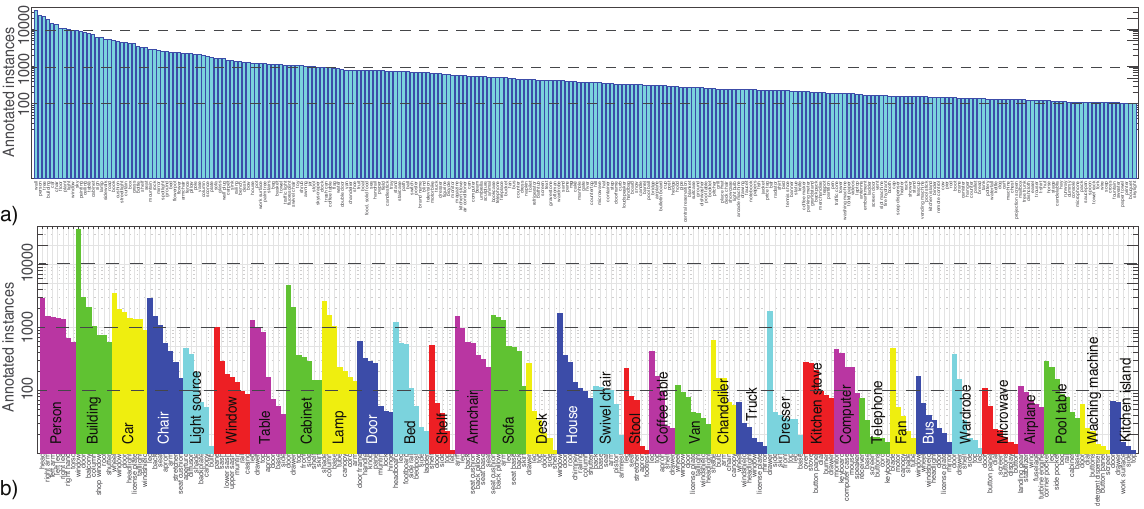}\\[1mm]
\includegraphics[width=1\textwidth]{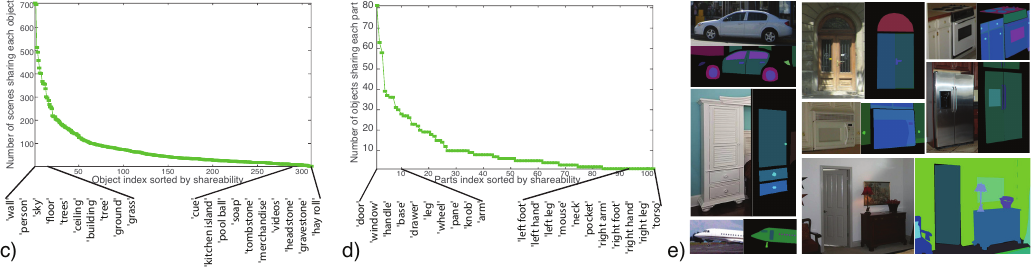}
\caption{a) Object classes sorted by frequency. Only the top 270 classes with more than 100 annotated instances are shown. 68 classes have more than a 1000 segmented instances. b) Frequency of parts grouped by objects. There are more than 200 object classes with annotated parts. Only objects with 5 or more parts are shown in this plot (we show at most 7 parts for each object class).  c) Objects ranked by the number of scenes they are part of. d) Object parts ranked by the number of objects they are part of. e) Examples of objects with doors. The bottom-right image is an example where the door does not behave as a part. }
\label{fig:stats}
\vspace{3mm}
\end{figure*}

Fig.~\ref{fig:stats}.a shows the distribution of ranked object frequencies. The distribution is similar to a Zipf's law and is typically found when objects are exhaustively annotated in images \citep{spainPerona10,xiao2010sun}. They differ from the ones from datasets such as COCO or ImageNet where the distribution is more uniform resulting from manual balancing.

Fig.~\ref{fig:stats}.b shows the distributions of annotated parts grouped by the objects to which they belong and sorted by frequency within each object class. Most object classes also have a non-uniform distribution of part counts. Fig.~\ref{fig:stats}.c and Fig.~\ref{fig:stats}.d show how objects are shared across scenes and how parts are shared by objects. Fig.~\ref{fig:stats}.e shows the variability in the appearances of the part `door'.

\begin{figure*}
\centering
\includegraphics[width=1\textwidth]{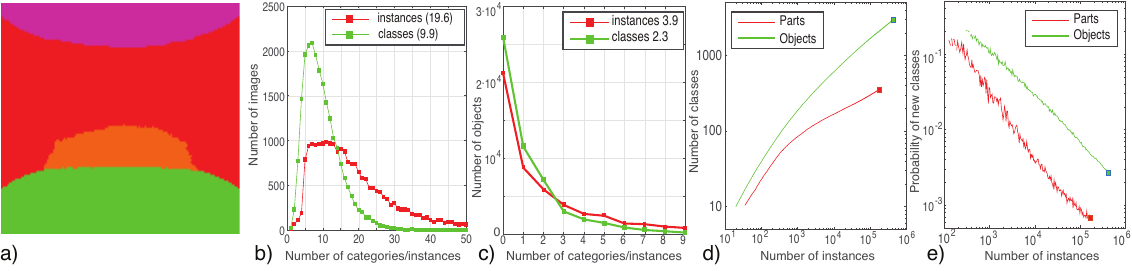}
\vspace{-3mm}
\caption{a) Mode of the object segmentations contains `sky', `wall', `building' and `floor'. b) Histogram of the number of segmented object instances and classes per image. c) Histogram of the number of segmented part instances and classes per object. d) Number of classes as a function of segmented instances (objects and parts). The squares represent the current state of the dataset. e) Probability of seeing a new object (or part) class as a function of the number of instances.}
\label{fig:stats2}
\end{figure*}

The mode of the object segmentations is shown in Fig.~\ref{fig:stats2}.a and contains the four objects (from top to bottom): `sky', `wall', `building' and `floor'. When using simply the mode to segment the images, it gets, on average, 20.9$\%$ of the pixels of each image right. Fig.~\ref{fig:stats2}.b shows the distribution of images according to the number of distinct classes and instances. On average there are 19.5 instances and 10.5 object classes per image, larger than other existing datasets (see Table~\ref{table:headings}). Fig.~\ref{fig:stats2}.c shows the distribution of parts.

As the list of object classes is not predefined, there are new classes appearing over time of annotation. Fig.~\ref{fig:stats2}.d shows the number of object (and part) classes as the number of annotated instances increases. Fig.~\ref{fig:stats2}.e shows the probability that instance $n+1$ is a new class after labeling $n$ instances. The more segments we have, the smaller the probability that we will see a new class. At the current state of the dataset, we get one new object class every 300 segmented instances. 

\subsection{Object-part relationships}
We analyze the relationships between the objects and object parts annotated in ADE20K. In the dataset, 76\% of the object instances have associated object parts, with an average of 3 parts per object. The class with the most parts is \textit{building}, with 79 different parts. On average, 10\% of the pixels correspond to object parts. A subset of the relation tree between objects and parts can be seen in Fig.~\ref{fig:tree_fig}. 

The information about objects and their parts provides interesting insights. For instance, we can measure in what proportion one object is part of another to reason about how strongly tied these are. For the object \textit{tree}, the most common parts are \textit{trunk} or \textit{branch}, whereas the least common are \textit{fruit}, \textit{flower} or \textit{leaves}.  

The object-part relationships can also be used to measure similarities among objects and parts, providing information about objects tending to appear together or sharing similar affordances. We measure the similarity between two parts as the common objects each one is part of. The most similar part to \textit{knob} is \textit{handle}, sharing objects such as \textit{drawer}, \textit{door} or \textit{desk}. Objects can similarly be measured by the parts they have in common. As such, \textit{chair's} most similar objects are \textit{armchair}, \textit{sofa} or \textit{stool}, sharing parts such as \textit{rail}, \textit{leg} or \textit{seat base}.

%, likewise the most similar parts to `propeller` are cockpit or landing gear and microphone, sharing . %

\subsection{Comparison with other datasets}
\label{sec:datasetcomparison}

We compare ADE20K with existing datasets in Tab.~\ref{table:headings}. Compared to the largest annotated datasets, COCO~\citep{lin2014microsoft} and Imagenet~\citep{ILSVRC15}, our dataset comprises of much more diverse scenes, where the average number of object classes per image is 3 and 6 times larger, respectively. With respect to SUN~\citep{xiao2010sun}, ADE20K is roughly 35\% larger  in terms of images and object instances. However, the annotations in our dataset are much richer since they also include segmentation at the part level. Such annotation is only available for the Pascal-Context/Part dataset~\citep{mottaghi2014role,chen2014parts} which contains 40 distinct part classes across 20 object classes. Note that we merged some of their part classes to be consistent with our labeling (e.g., we mark both \emph{left leg} and \emph{right leg} as the same semantic part \emph{leg}). Since our dataset contains part annotations for a much wider set of object classes, the number of part classes is almost 9 times larger in our dataset. 

An interesting fact is that any image in ADE20K contains at least 5 objects, and the maximum number of object instances per image reaches 273, and 419 instances, when counting parts as well. This shows the high annotation complexity of our dataset.

\begin{table*}[t!]
\begin{center}
\caption{Comparison with existing datasets with semantic segmentation.}
\label{table:headings}
\begin{tabular}{lcccccc}
\hline\noalign{\smallskip}
  & Images & Obj. inst. & Obj. classes  & Part inst.  & Part classes & Obj. classes per image \\
\noalign{\smallskip}
\hline
\noalign{\smallskip}
COCO  & 123,287 & 886,284 & 91 & 0 & 0 & 3.5\\
ImageNet$^*$ & 476,688 & 534,309 & 200 & 0 & 0 & 1.7\\
NYU Depth V2 & 1,449 & 34,064 & 894 & 0 & 0 & 14.1 \\
Cityscapes & 25,000 & 65,385 & 30 & 0 & 0 & 12.2 \\
SUN & 16,873 & 313,884 & 4,479 & 0 & 0 & 9.8\\
OpenSurfaces & 22,214 & 71,460 & 160 & 0 & 0 & N/A\\
PascalContext & 10,103 & $\sim$104,398$^{**}$ & 540 & 181,770 & 40 & 5.1\\
ADE20K & 22,210 & 434,826 & 2,693 & 175,961 & 476 & 9.9\\
\hline
\end{tabular}
\end{center}
\vspace{-3mm}
\scriptsize $^*$ has only bounding boxes (no pixel-level segmentation). Sparse annotations.\\
\scriptsize $^{**}$ PascalContext dataset does not have instance segmentation. In order to estimate the number of instances, we find connected components (having at least 150pixels) for each class label. 
\end{table*}
\setlength{\tabcolsep}{1.4pt}

\section{Pixel-wise Scene Understanding Benchmarks}
\label{section:benchmarks}
Based on the data of the ADE20K, we construct two benchmarks for pixel-wise scene understanding: scene parsing and instance segmentation: 

\begin{itemize}
\item \textbf{Scene parsing.} Scene parsing is to segment the whole image densely into semantic classes, where each pixel is assigned a class label such as the region of \textit{tree} and the region of \textit{building}. 
%\item \textbf{Part segmentation.} Part segmentation is to segment the the parts of pre-defined list of object categories. 
\item \textbf{Instance segmentation.} Instance segmentation is to detect the object instances inside an image and further generate the precise segmentation masks of the objects. Its difference compared to the task of scene parsing is that in scene parsing there is no instance concept for the segmented regions, instead in instance segmentation if there are three persons in the scene, the network is required to segment each one of the person regions.
\end{itemize}

We introduce the details of each task and the baseline models we train as below.

\subsection{Scene parsing benchmark}

We select the top 150 categories ranked by their total pixel ratios\footnote{As the original images in the ADE20K dataset have various sizes, for simplicity we rescale those large-sized images to make their minimum heights or widths as 512 in the SceneParse150 benchmark.} in the ADE20K dataset and build a scene parsing benchmark of ADE20K, termed as \textit{SceneParse150}. Among the 150 categories, there are 35 stuff classes (\textit{i.e., wall, sky, road}) and 115 discrete object classes (\textit{i.e., car, person, table}). The annotated pixels of the 150 classes occupy 92.75\% of all the pixels of the dataset, where the stuff classes occupy 60.92\%, and discrete object classes occupy 31.83\%.

\begin{figure*}
\begin{center}
\includegraphics[width=1\textwidth]{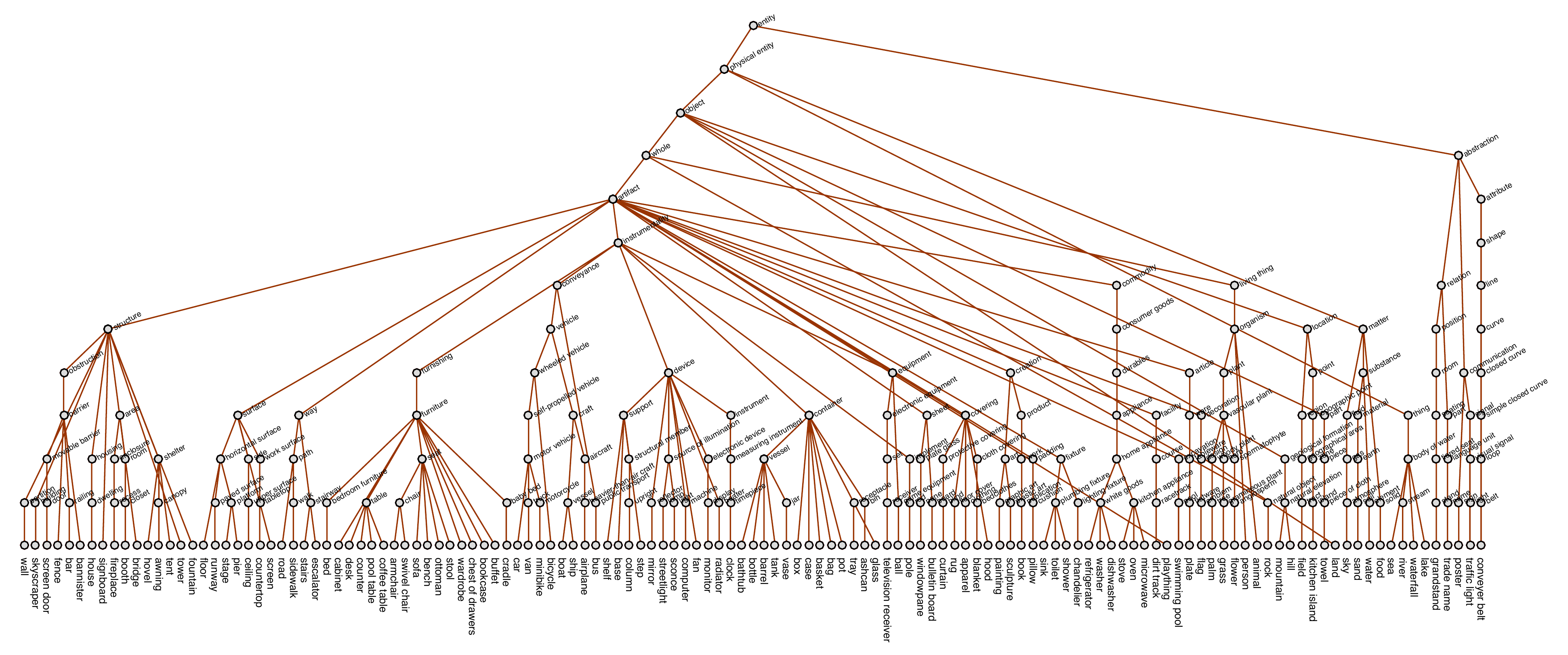}
\end{center}
\vspace{-5mm}
\caption{Wordnet tree constructed from the 150 objects in the SceneParse150 benchmark. Clusters inside the wordnet tree represent various hierarchical semantic relations among objects.}
\label{wordnet_tree}
\end{figure*}

We map the WordNet synsets with each one of the object names, then build up a WordNet tree through the hypernym relations of the 150 categories shown in Fig.~\ref{wordnet_tree}. We can see that these objects form several semantic clusters in the tree, such as the \textit{furniture} synset node containing \textit{cabinet}, \textit{desk}, \textit{pool table}, and \textit{bench}, the \textit{conveyance} node containing \textit{car}, \textit{truck}, \textit{boat}, and \textit{bus}, as well as the \textit{living thing} node containing \textit{shrub}, \textit{grass}, \textit{flower}, and \textit{person}. Thus, the structured object annotation given in the dataset bridge the image annotation to a wider knowledge base. 

As for baseline networks for scene parsing on our benchmark, we train several semantic segmentation networks: SegNet~\citep{badrinarayanan2015segnet}, FCN-8s~\citep{long2015fully}, DilatedVGG, DilatedResNet~\citep{CP2016Deeplab, YuKoltun2016}, two cascade networks proposed in \cite{zhou2017scene} where the backbone models are SegNet and DilatedVGG. We train these models on NVIDIA Titan X GPUs. 

%SegNet has encoder and decoder architecture for image segmentation; FCN upsamples the activations of multiple layers in the CNN for pixelwise segmentation; 

%We integrate the proposed cascade segmentation module on the two baseline networks: SegNet and DilatedNet. We did not integrate it with FCN because the original FCN requires a large amount of GPU memory and has skip connections across layers. For the Cascade-SegNet, two streams share a single encoder, from \texttt{conv1\_1} to \texttt{conv5\_3}, while each stream has its own decoder, from \texttt{deconv5\_3} to \texttt{loss}. For the Cascade-DilatedNet, the two streams split after \texttt{pool3}, and keep spatial dimensions of their feature maps afterwards. For a fair comparison and benchmark purposes, the cascade networks only have stuff stream and object stream. We train these network models using the Caffe library~\citep{jia2014caffe} on NVIDIA Titan X GPUs.

Results are reported in four metrics commonly used for semantic segmentation~\citep{long2015fully}: 
\begin{itemize}
\item \textbf{Pixel accuracy} indicates the proportion of correctly classified pixels;
\item \textbf{Mean accuracy} indicates the proportion of correctly classified pixels averaged over all the classes. 
\item \textbf{Mean IoU} indicates the intersection-over-union between the predicted and ground-truth pixels, averaged over all the classes. 
\item \textbf{Weighted IoU} indicates the IoU weighted by the total pixel ratio of each class.
\end{itemize}

Since some classes like \textit{wall} and \textit{floor} occupy far more pixels of the images, pixel accuracy is biased to reflect the accuracy over those few large classes. Instead, mean IoU reflects how accurately the model classifies each discrete class in the benchmark. The scene parsing data and the development toolbox are released in the Scene Parsing Benchmark website\footnote{\url{http://sceneparsing.csail.mit.edu}}. 

\begin{table}
\caption{Baseline performance on the validation set of SceneParse150.}
\label{performance_segmentation}
\centering
\footnotesize
\addtolength{\tabcolsep}{0.8pt}
\begin{tabular}{ l | c | c | c | c }
  \hline                       
  Networks &Pixel Acc. & Mean Acc. & Mean IoU  & Weighted IoU\\
  \hline  
FCN-8s & 71.32\% & 40.32\% & 0.2939 & 0.5733 \\
SegNet & 71.00\% & 31.14\% & 0.2164 & 0.5384 \\
DilatedVGG & 73.55\% & 44.59\% & 0.3231 & 0.6014 \\
DilatedResNet-34 & 76.47\% & 45.84\% & 0.3277 & 0.6068\\
DilatedResNet-50 & 76.40\% & 45.93\% & 0.3385 & 0.6100 \\
Cascade-SegNet & 71.83\% & 37.90\% & 0.2751 & 0.5805 \\
Cascade-DilatedVGG & 74.52\% & 45.38\% & 0.3490 & 0.6108 \\
\hline
\end{tabular}
\end{table}

The segmentation performance of the baseline networks on SceneParse150 is listed in Table~\ref{performance_segmentation}. Among the baselines, the networks based on dilated convolutions achieve better results in general than FCN and SegNet. Using the cascade framework, the performance further improves. In terms of mean IoU, Cascade-SegNet and Cascade-DilatedVGG outperform SegNet and DilatedVGG by 6\% and 2.5\%, respectively.

%We further decompose the performance of networks on 35 stuff and 115 discrete object classes respectively, in Table~\ref{performance_object115stuff35}. We observe that the two cascade networks perform much better on the 115 discrete objects compared to the baselines. This validates that the design of cascade module helps improve scene parsing for the discrete objects as they have less training data but more visual complexity compared to those stuff classes. %For stuff, Cascade-SegNet achieves a lower accuracy and a slightly higher IoU, showing that it is more tuned to segment the objects rather than the stuff classes.

\begin{figure*}
\begin{center}
\includegraphics[clip, trim=0cm 3.4cm 0cm 0cm, width=0.95\textwidth]{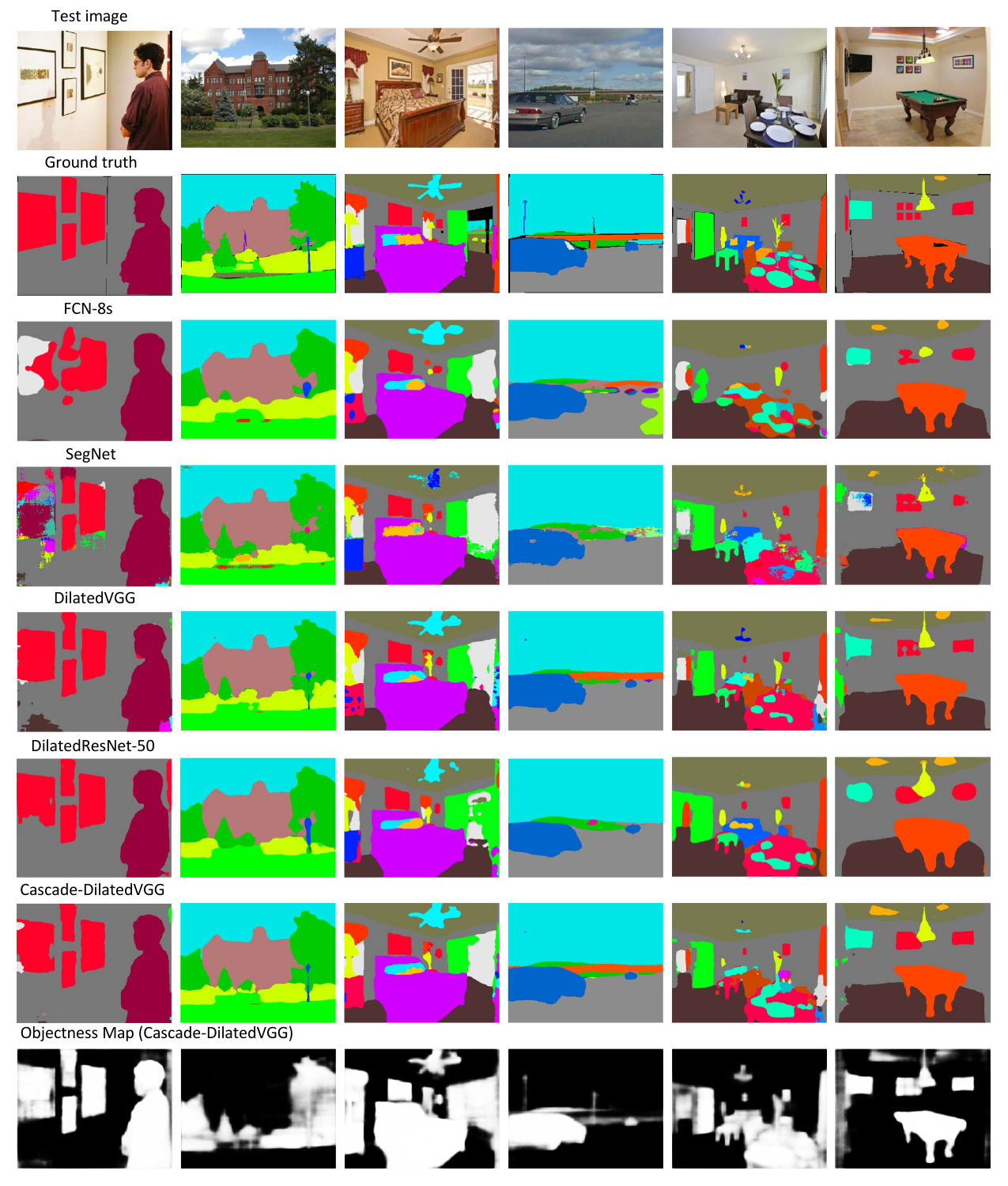}
\end{center}
\caption{Ground-truths, scene parsing results given by the baseline networks. All networks can give correct predictions for the common, large object and stuff classes, the difference in performance comes mostly from small, infrequent objects and how well they handle details.}
\label{segmentation-result}
\end{figure*}

\begin{figure*}
\begin{center}
\includegraphics[width=1\textwidth]{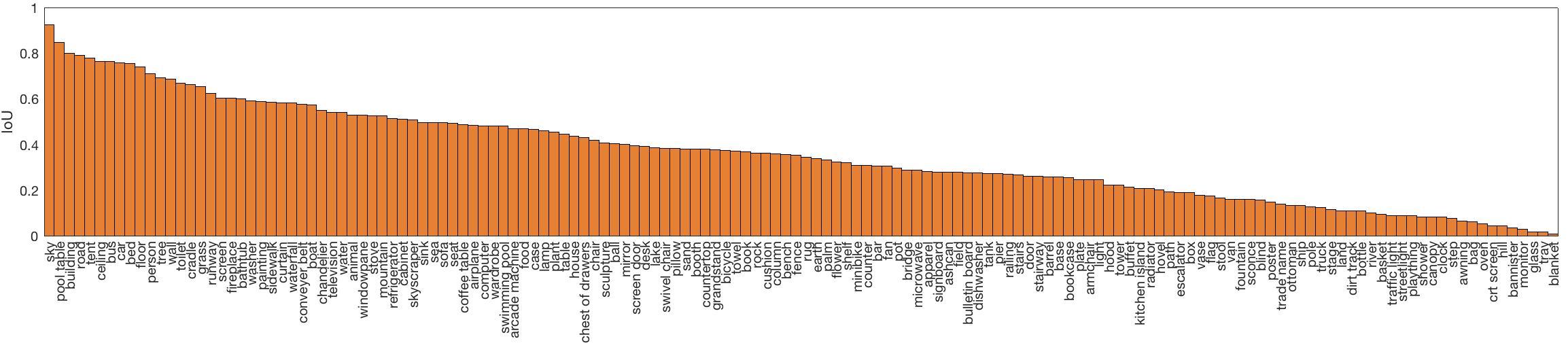}
\end{center}
\caption{Plot of scene parsing performance (IoU) on the 150 categories achieved by DilatedResNet-50 model. The best segmented categories are stuff, and the worst segmented categories are objects that are usually small and have few pixels.}
\label{plot_iou}
\end{figure*}

Qualitative scene parsing results from the validation set are shown in Fig.~\ref{segmentation-result}. We observe that all the baseline networks can give correct predictions for the common, large object and stuff classes, the difference in performance comes mostly from small, infrequent objects and how well they handle details. We further plot the IoU performance of all the 150 categories given by the baseline model DilatedResNet-50 in Fig.~\ref{plot_iou}. We can see that the best segmented categories are stuffs like \textit{sky}, \textit{building} and \textit{road}; the worst segmented categories are objects that are usually small and have few pixels, like \textit{blanket}, \textit{tray} and \textit{glass}.

\subsection{Opening source the state-of-the-art scene parsing models}
Since the introduction of SceneParse150 firstly in 2016, it has become a standard benchmark for evaluating new semantic segmentation models. However, the state-of-the-art models are in different libraries (Caffe, PyTorch, Tensorflow) while training codes of some models are not released, which makes it hard to reproduce the original results reported in the paper. To benefit the research community, we re-implement several state-of-the-art models in PyTorch and open source them\footnote{Reimplementation of the state-of-the-art models are released at \url{https://github.com/CSAILVision/semantic-segmentation-pytorch}}. Particularly, we implement (1) The plain dilated segmentation network which use the dilated convolution \cite{YuKoltun2016}; (2) PSPNet proposed in \cite{zhao2017pyramid}, it introduces Pyramid Pooling Module (PPM) to aggregate multi-scale contextual information in the scene; (3) UPerNet proposed in \cite{xiao2018unified} which adopts architecture like Feature Pyramid Network (FPN) \cite{lin2017feature} to incorporate multi-scale context more efficiently. Table \ref{performance_segmentation_sota} shows results on the validation set of SceneParse150. Compared to plain DilatedResNet, PPM and UPerNet architectures improve mean IoU by 3-7\%, and pixel accuracy by 1-2\%. The superior performance shows the importance of context in the scene parsing task.

\begin{table}
\caption{Reimplementation of state-of-the art models on the validation set of SceneParse150. PPM refers to Pyramid Pooling Module.}
\label{performance_segmentation_sota}
\centering
\footnotesize
\addtolength{\tabcolsep}{0.8pt}
\begin{tabular}{ l | c | c }
  \hline                       
  Networks &Pixel Acc.  & Mean IoU  \\
  \hline  
DilatedResNet-18 & 77.41\% & 0.3534 \\
DilatedResNet-50 & 77.53\% & 0.3549 \\
DilatedResNet-18 + PPM \cite{zhao2017pyramid}& 78.64\% & 0.3800 \\
DilatedResNet-50 + PPM \cite{zhao2017pyramid}& 80.23\% & 0.4204 \\
DilatedResNet-101 + PPM  \cite{zhao2017pyramid}& 80.91\% & 0.4253\\
UPerNet-50 \cite{xiao2018unified}& 80.23\% & 0.4155  \\
UPerNet-101 \cite{xiao2018unified}& 81.01\% & 0.4266  \\

\hline
\end{tabular}
\end{table}

% \begin{table}
% \caption{Performance of stuff and discrete object segmentation.}
% \label{performance_object115stuff35}
% \footnotesize
% \centering
% \addtolength{\tabcolsep}{1.7pt}
% \begin{tabular}{| l | c | c | c | c |}
%   \hline  
%   & \multicolumn{2}{|c|}{35 stuff} & \multicolumn{2}{|c|}{115 discrete objects} \\ 
%   \hline                       
%   Networks & Mean Acc. & Mean IoU & Mean Acc. &  Mean IoU \\
%     \hline  
% FCN-8s & 46.74\% & 0.3344 & 38.36\% & 0.2816 \\
% SegNet & 43.17\% & 0.3051 & 27.48\% & 0.1894 \\
% DilatedNet & 49.03\% & 0.3729 & 43.24\% & 0.3080 \\
% Cascade-SegNet & 40.46\% & 0.3245 & 37.12\% & 0.2600 \\
% Cascade-DilatedNet & \textbf{49.80}\% & \textbf{0.3779} & \textbf{44.04}\% & \textbf{0.3401} \\
%     \hline
% \end{tabular}
% \end{table}

% \begin{figure*}
% \begin{center}
% \includegraphics[width=1\textwidth]{figures/comparison_result_plot.pdf}
% \end{center}
% \caption{The IoU of each class for the Cascade-SegNet, SegNet, and FCN on semantic segmentation of scenes. The x-axis is ranked by the IoUs of the Cascade-SegNet.}
% \label{result_IOUclass}
% \end{figure*}

\subsection{Effect of batch normalization for scene parsing}
\label{semantic_segm_analysis}
An overwhelming majority of semantic segmentation models are fine-tuned from a network trained on ImageNet~\cite{ILSVRC15}, the same as most of the object detection models~\cite{ren2015faster,lin2017feature,he2017mask}. There has been work~\cite{peng2018megdet} exploring the effects of the size of batch normalization (BN)~\cite{ioffe2015batch}. The authors discovered that, if a network is trained with BN, only by a sufficiently large batch size of BN can the network achieves state-of-the-art performance. We conduct control experiments on ADE20K to explore the issue in terms of semantic segmentation. Our experiment shows that a reasonably large batch size is essential for matching the highest score of the-state-or-the-art models, while a small batch size such as 2 in Table ~\ref{batch_norm_analysis} lower the score of the model significantly by 5\%. Thus training with a single GPU with limited RAM or with multiple GPUs under unsynchronized BN is unable to reproduce the best reported numbers. The possible reason is that the BN statics, i.e., mean and standard variance of activations may not be accurate when batch size is not sufficient. 

\begin{table}
\caption{Comparisons of models trained with various batch normalization settings. The framework used is a Dilated ResNet-50 with Pyramid Pooling Module.}
\label{batch_norm_analysis}
\centering
\footnotesize
\addtolength{\tabcolsep}{0.8pt}
\begin{tabular}{ l | c | c | c | c }
  \hline                       
  BN Status & Batch Size & BN Size & Pixel Acc. & Mean IoU  \\
  \hline  
Synchronized & 16 & 16 & 79.73\% & 0.4126 \\
& 8 & 8 & 80.05\% & 0.4158 \\
& 4 & 4 & 79.71\% & 0.4119 \\
& 2 & 2 & 75.26\% & 0.3355 \\
\hline 
Unsynchronized & 16 & 2 & 75.28\% & 0.3403 \\
\hline
Frozen & 16 & N/A & 78.32\% & 0.3809 \\
& 8 & N/A & 78.29\% & 0.3793 \\
& 4 & N/A & 78.34\% & 0.3833 \\
& 2 & N/A & 78.81\% & 0.3856 \\

\hline
\end{tabular}
\end{table}

Our baseline framework is the PSPNet with a dilated ResNet-50 as the backbone. Besides those BN layers in the ResNet, they are also used in the PPM. The baseline framework is trained with 8 GPUs and 2 images on each GPU. We adopt synchronized BN for the baseline network, \emph{i.e.}, the BN size should be the same as the batch size. Besides the synchronized BN setting, we also have unsynchronized BN setting and frozen BN setting. The former one means that the BN size is the number of images on each GPU; the latter one means that the BN layers are frozen in the backbone network, and removed from the PPM. The training iterations and learning rate are set to $100k$ and $0.02$ for the baseline, respectively. For networks trained under the frozen BN setting, the learning rate for the network with 16 batch size is set to $0.004$ to prevent gradient explosion. And for networks with batch size smaller than 16, we both linearly decrease the learning rate and increase the training iterations according to previous works~\cite{goyal2017accurate}. Different from Table~\ref{performance_segmentation_sota}, the results are obtained w/o multi-scale testing.

We report the results in Table~\ref{batch_norm_analysis}. In general, we empirically find that using BN layers with a sufficient BN size leads to better performance. The model with batch size and BN size as 16 (line 2) outperforms the one with batch size 16 and frozen BN (line 7) by $1.41\%$ and $3.17\%$ in terms of Pixel Acc. and Mean IoU respectively. We witness negligible changes of performance when batch (and BN) size changes in the range from 4 to 16 under synchronized BN setting (line 2-4). However, when the BN size drops to 2, the performance downgrades significantly (line 5). Thus a BN size of 4 is the inflection point in our experiments. This finding is different from the finding for object detection~\cite{peng2018megdet}, in which the inflection point is at a BN size of 16. We conjecture that it is due to images for semantic segmentation are densely annotated, different from those for object detection with bounding-box annotations. Therefore it is easier for semantic segmentation networks to obtain more accurate BN statistics with fewer images.

When we experiment with unsynchronized BN setting, \emph{i.e.}, we increase the batch size but do not change the BN size (line 6), the model yields almost identical result compared with the one with the same BN size but smaller batch size (line 5). Also, when we freeze the BN layers during the fine-tuning, the models are not sensitive to the batch size (line 7-10). These two set of experiments indicate that, for semantic segmentation models, the BN size is the one that matters instead of the batch size. But we do note that smaller batch size leads to longer training time because we need to increase the training iterations for models with small batch size.

\subsection{Instance Segmentation}
\label{instance_segm_sec}
\begin{figure*}
\begin{center}
\includegraphics[width=1.0\textwidth]{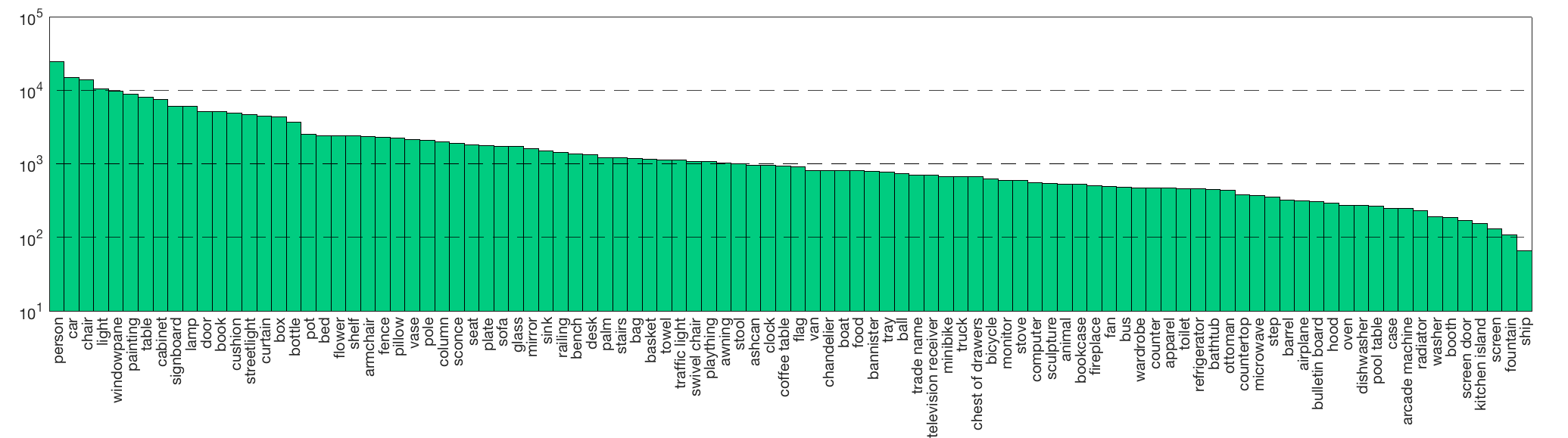}
\end{center}
\caption{Instance number per object in instance segmentation benchmark. All the objects except \textit{ship} have more than 100 instances.}
\label{instance_number}
\end{figure*}

To benchmark the performance of instance segmentation, we select 100 foreground object categories from the full dataset, term it as InstSeg100. The plot of the instance number per object in InstSeg100 is shown in Fig.~\ref{instance_number}. The total number of object instances is 218K, on average there are 2.2K instances per object category and 10 instances per image; all the objects except \textit{ship} have more than 100 instances.

\begin{figure*}
\begin{center}
\includegraphics[width=0.95\textwidth]{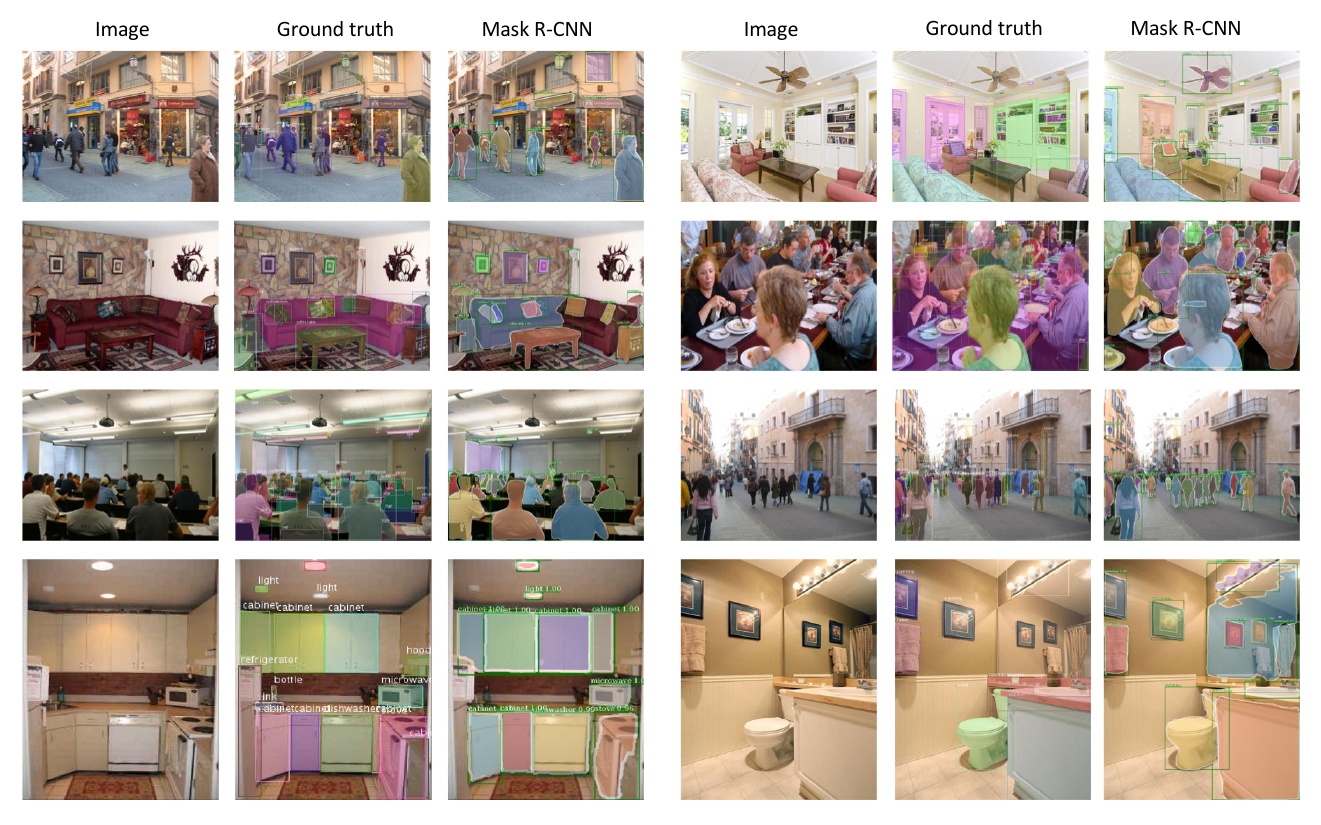}
\end{center}
\caption{Images, ground-truths, and instance segmentation results given by multi-scale Mask R-CNN model.}
\label{instancesegmentation-maskRCNN}
\end{figure*}

We use Mask R-CNN~\cite{he2017mask} models as baselines for InstSeg100. The models use FPN-50 as backbone network, initialized from ImageNet, other hyper-parameters strictly follow those used in~\cite{he2017mask}. Two variants are presented, one with single scale training, the other with multi-scale training, their performance on the validation set is shown in Table~\ref{table_instancesegmentation}. We report an overall metric mean Average Precision mAP, along with metrics on different object scales, denoted by mAP$_{S}$ (objects smaller than $32\times32$ pixels), mAP$_{M}$ (between $32\times32$ and $96\times96$ pixels) and mAP$_{L}$ (larger than $96\times96$ pixels). Numbers suggest that (1) multi-scale training could greatly improve the average performance ($\sim0.04$ in mAP); (2) instance segmentation of small objects on our dataset is extremely challenging, it does not improve ($\sim0.02$) as much as large objects ($\sim0.07$) when using multi-scale training.
Qualitative results of the Mask R-CNN model are presented in Fig.~\ref{instancesegmentation-maskRCNN}. We can see that it is a strong baseline, giving correct detections and accurate object boundaries. Some typical errors are object reflections in the mirror, as shown in the bottom right example.

\begin{table}
\begin{center}
\caption{Baseline performance on the validation set of InstSeg100.}\label{table_instancesegmentation}
\begin{tabular}{ l | c c c | c}
  	\hline                       
  	Networks & mAP$_{S}$ & mAP$_{M}$ & mAP$_{L}$ & mAP \\ \hline  
	Mask R-CNN single-scale & .0542 & .1737 & .2883 & .1832 \\
	Mask R-CNN multi-scale & .0733 & .2256 & .3584 & .2241 \\
\hline
\end{tabular}
\end{center}

\end{table}

% --------------------------------------------------
\begin{figure*}
\begin{center}
\includegraphics[width=1.\textwidth, trim={50, 0, 50, 0}, clip]{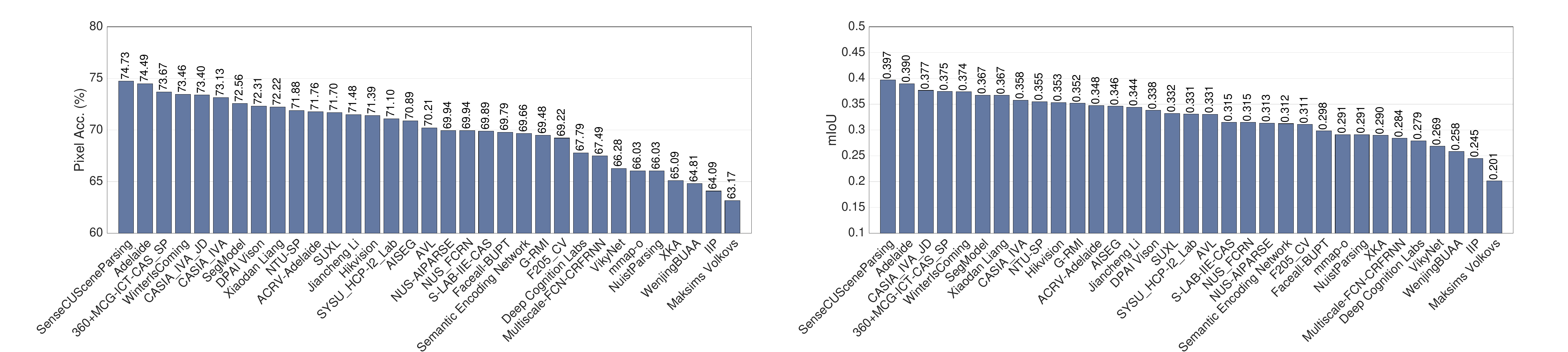}
\end{center}
\caption{Scene Parsing Track Results, ranked by pixel accuracy and mean IoU.} 
%b) Pixel accuracy for the top scene parsing submissions, semantic segmentation baseline models and human agreement. }
\label{challenge_sceneparsing}
\end{figure*}

\begin{table}
\caption{Scene parsing performance before and after fusing outputs from instance segmentation model Mask R-CNN.}
\label{tab:instance_help_semantic}
\centering
\footnotesize
\addtolength{\tabcolsep}{0.8pt}
\begin{tabular}{ l | c | c | c | c}
  \hline                       
  Networks & \multicolumn{2}{c}{Pixel Acc.} & \multicolumn{2}{c}{Mean IoU} \\
   & Before & After & Before & After \\
  \hline  
DilatedResNet-50 + PPM \cite{zhao2017pyramid} & 80.23\% & 80.21\% & 0.4204 & 0.4256 \\
DilatedResNet-101 + PPM \cite{zhao2017pyramid} & 80.91\% & 80.91\% & 0.4253 & 0.4290 \\

\hline
\end{tabular}
\end{table}

\subsection{How does scene parsing performance improve with instance information?}
In the previous sections, we train and test semantic and instance segmentation tasks separately. Given that instance segmentation is trained with additional instance information compared to scene parsing, we further analyze how instance information can assist scene parsing. 

Instead of re-modeling, we study this problem by fusing results from our trained state-of-the-art models, PSPNet for scene parsing and Mask R-CNN for instance segmentation. Concretely, we first take Mask R-CNN outputs and threshold predicted instances by confidence ($\geq0.95$); then we overlay the instance masks on to the PSPNet predictions; if one pixel belongs to multiple instances, it takes the semantic label with the highest confidence. Note that instance segmentation only works for 100 foreground object categories as opposed to 150 categories, so stuff predictions come from the scene parsing model. Quantitative results are shown in \ref{tab:instance_help_semantic}, overall the fusion improves scene parsing performance, pixel accuracy stays around the same and mean IoU improves around 0.4-0.5\%. This experiment demonstrate that instance level information is useful for helping the non-instance-aware scene parsing task.

\section{Places Challenges}
\label{section:challenges}
In order to foster new models for pixel-wise scene understanding, we organized in 2016 and 2017 the Places Challenge including the scene parsing track and instance segmentation track. %Scene parsing submissions were ranked based on the average score of the mean IoU and pixel-wise accuracy in the benchmark test set. For instance segmentation, we used the mean Average Precision (mAP), following COCO's evaluation metrics.

\subsection{Scene Parsing Track}
\begin{table}
\caption{Top performing models in Scene Parsing for Places Challenge 2016.}
\label{performance_segmentationchallenge2016}
\centering
\footnotesize
\addtolength{\tabcolsep}{0.8pt}
\begin{tabular}{l | c | c | c }
  \hline                       
  Team &Pixel Acc. & Mean IoU  & Score\\
  \hline  
SenseCUSceneParsing \cite{zhao2017pyramid} & 74.73\% & .3968 & .5720  \\
Adelaide \cite{widerordeepersceneparsing} & 74.49 \% & .3898  & .5673  \\
360-MCG-CT-CAS\_SP & 73.67\% & .3746 & .5556 \\

\hline
\end{tabular}
\end{table}

\begin{table}
\caption{Top performing models in Scene Parsing for Places Challenge 2017.}
%\textcolor{red}{is it possible to find the reference for the methods?}}
\label{performance_segmentationchallenge2017}
\centering
\footnotesize
\addtolength{\tabcolsep}{0.8pt}
\begin{tabular}{ l | c | c | c }
  \hline                       
  Team &Pixel Acc. & Mean IoU  & Score\\
  \hline  
CASIA\_IVA\_JD & 73.40\% &.3754 & .5547  \\
WinterIsComing & 73.46\% & .3741 & .5543  \\
Xiaodan Liang & 72.22\% & .3672 & .5447 \\
\hline
\end{tabular}
\end{table}

Scene parsing submissions were ranked based on the average score of the mean IoU and pixel-wise accuracy in the benchmark test set. 

The Scene Parsing Track totally received 75 submissions from 22 teams in 2016 and 27 submissions from 11 teams in 2017. The top performing teams for both years are shown in Table \ref{performance_segmentationchallenge2016} and Table \ref{performance_segmentationchallenge2017}. The winning team in 2016 proposing PSPNet \cite{zhao2017pyramid} still holds the highest score. Fig.~\ref{qual_sceneparsing} shows some qualitative results from the top performing models on each year. 

\begin{figure}
\begin{center}
\includegraphics[width=0.5\textwidth, trim={30 0 0 0}, clip]{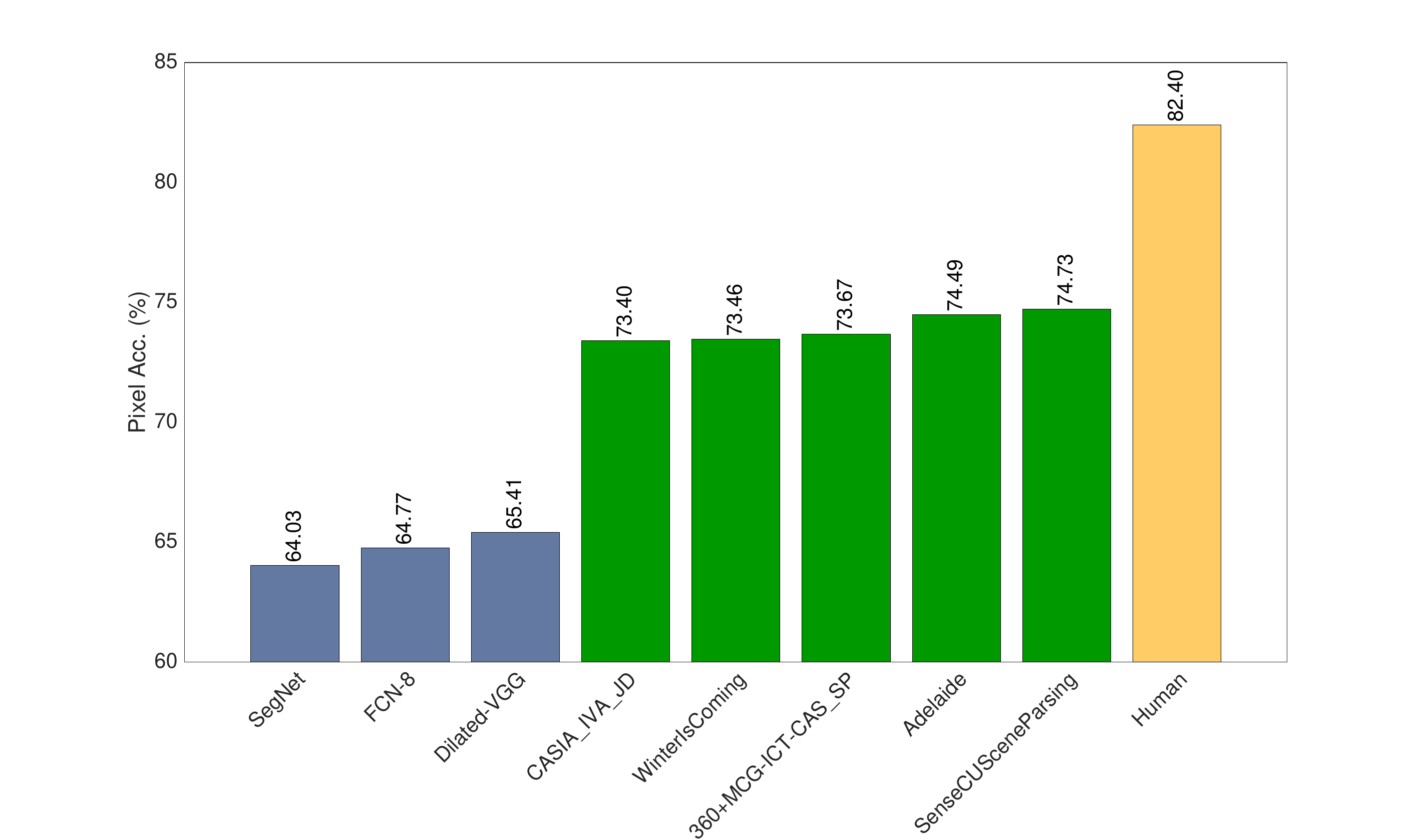}
\end{center}
\caption{Top scene parsing models compared with human performance and baselines in terms of pixel accuracy. Scene parsing based on the image mode has a 20.30\% pixel accuracy.}
\label{semantic_segm_compare}
\end{figure}

\begin{figure*}
\begin{center}
\includegraphics[width=1\textwidth, trim={0 0 0 0}, clip]{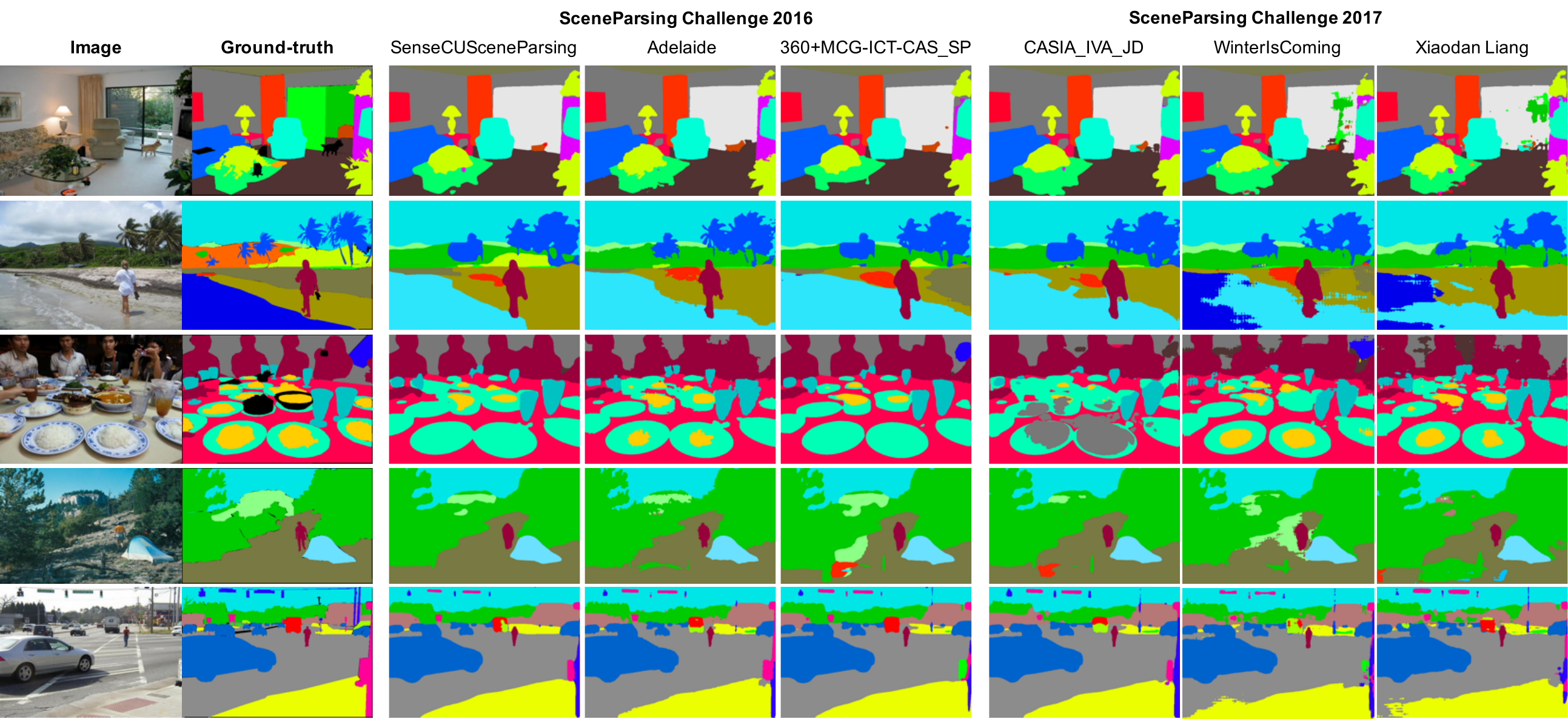}
\end{center}
\caption{Scene Parsing results given by top methods for Places Challenge 2016 and 2017.}
\label{qual_sceneparsing}
\end{figure*}

In Fig.~\ref{semantic_segm_compare} we compare the top models against the proposed baselines and human performance (approximately measured as the annotation consistency in Sec.\ref{sec:consistency}), which could be the upper bound performance. As an interesting comparison, if we use the image mode generated in Fig.\ref{fig:stats2} as prediction on the testing set, it achieves 20.30\% pixel accuracy, which could be the lower bound performance for all the models.

\begin{figure*}
\begin{center}
\includegraphics[width=1.\textwidth]{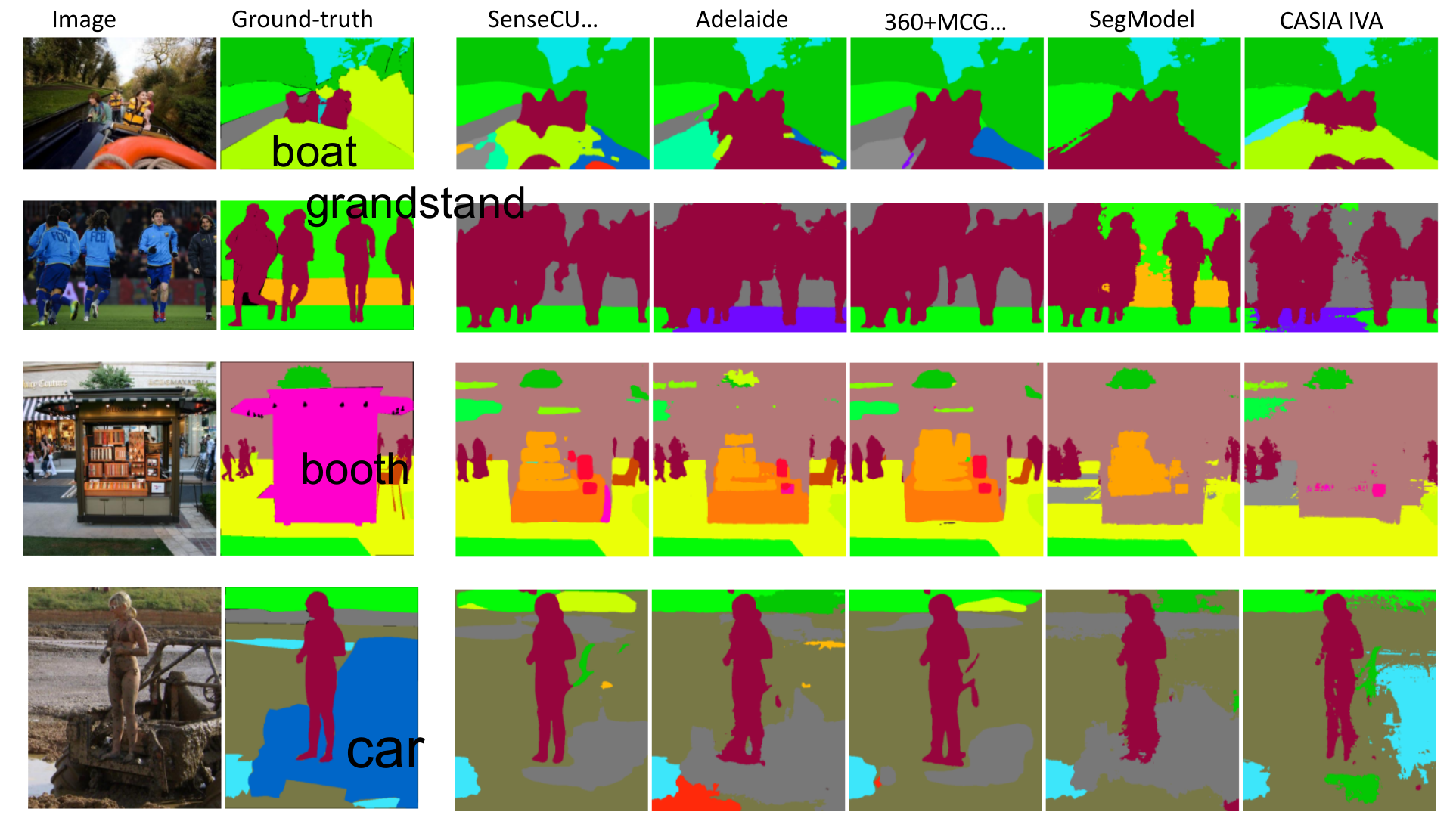}
\end{center}
\caption{Ground-truths and predictions given by top methods for scene parsing. The mistaken regions are labeled. We can see that models make mistakes on objects in non-canonical views such as the boat in first example, and on objects which require high-level reasoning such as the muddy car in the last example. 
%\textcolor{red}{This figure is a bit messy, numbers do not align, titles have ..., overlaid texts are sometimes italic.}
}
\label{errorcase_sceneparsing}
\end{figure*}

Some error cases are shown in Fig.~\ref{errorcase_sceneparsing}. We can see that models usually fail to detect the concepts in some images that have occlusions or require high-level context reasoning. For example, the boat in the first image is not a typical view of a boat so that the models fail; For the last image, the muddy car is missed by all the top performer networks because of its muddy camouflage. 

\subsection{Instance Segmentation Track}

For instance segmentation, we used the mean Average Precision (mAP), following COCO's evaluation metrics.

\begin{figure*}
\begin{center}
\includegraphics[width=1\textwidth, trim={0 0 0 0}, clip]{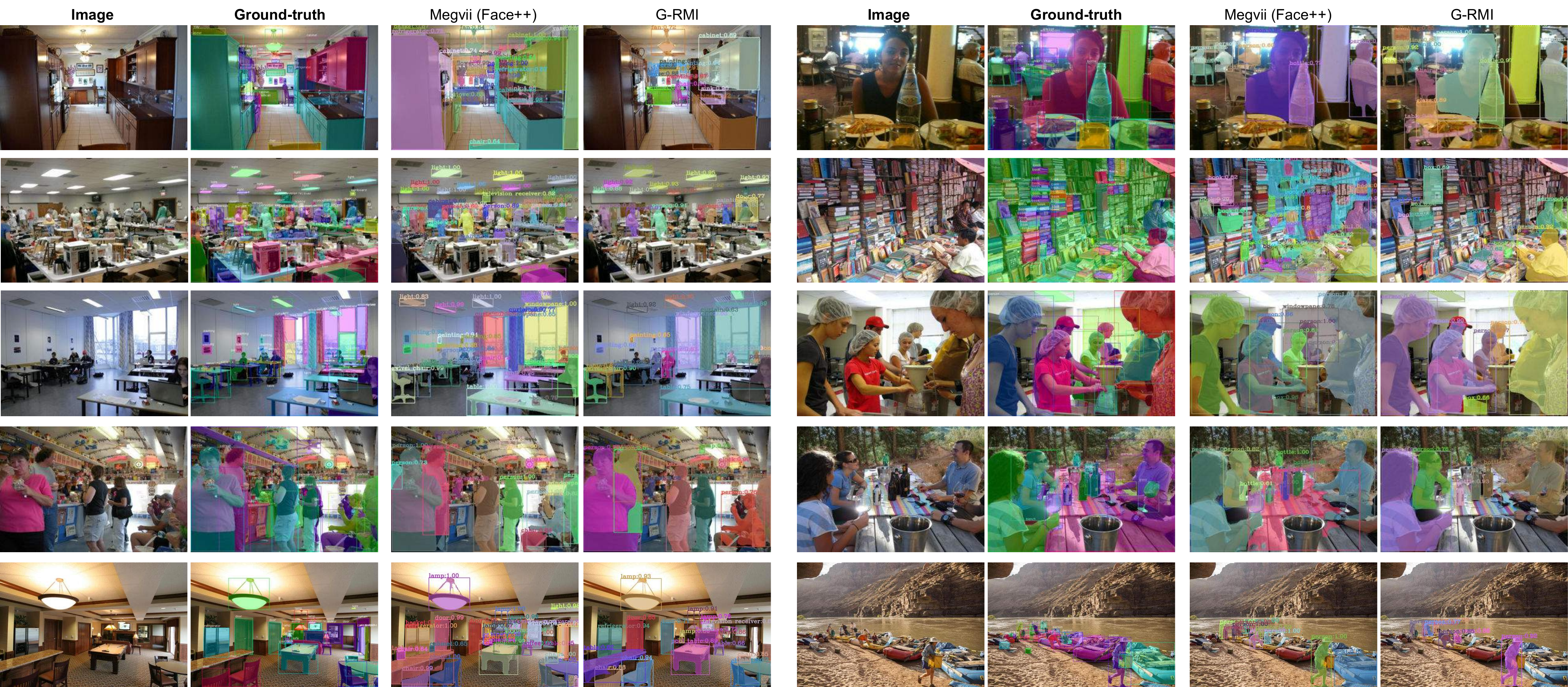}
\end{center}
\caption{Instance Segmentation results given by top methods for Places Challenge 2017.}
\label{qual_instance}
\end{figure*}

The Instance Segmentation Track, introduced in Places Challenge 2017, received 12 submissions from 5 teams. Two teams beat the strong Mask R-CNN baseline by a good margin, their best model performances are shown in Table \ref{performance_instancechallenge2017} together with the Mask R-CNN baseline trained by ourselves. The performances for small, medium and large objects are also reported, following \ref{instance_segm_sec}. Fig. \ref{qual_instance} shows qualitative results from the teams' best models.

\begin{table}
\caption{Top performing models in Instance Segmentation for Places Challenge 2017. }
\label{performance_instancechallenge2017}
\centering
\footnotesize
%\addtolength{\tabcolsep}{0.8pt}
\begin{tabular}{ l | c c c | c }
\hline                       
Team & mAP$_{S}$ & mAP$_{M}$ & mAP$_{L}$ & mAP\\
\hline  
Megvii (Face++) \cite{peng2018megdet} & .1386 & .3015 & .4119 & .2977    \\
G-RMI & .0980 & .2523 &  .3858 & .2415   \\
\hline
Baseline Mask R-CNN & .0733 & .2256 & .3584 & .2241 \\

\hline
\end{tabular}
\end{table}

As can be seen in table \ref{performance_instancechallenge2017}, both methods outperform the Mask R-CNN at any of the object scales, even though they still struggle with medium and small objects. Megvii (Face++) submission seems to particularly advantage G-RMI for the small objects, probably due to the use of contextual information. Their mAP on small objects show a relative improvement over G-RMI of 41\%, compared to the 19\% and 6\% of medium and large objects.  

This effect can be qualitatively seen in figure \ref{qual_instance}. While both methods perform similarly well in finding large object classes such as people or tables, Megvii (Face++) is able to detect small paintings (rows 1 and 3) or lights (row 5) occupying small regions.

\subsection{Take-aways from the Challenge}
Looking at the challenge results, there are several peculiarities that make ADE20K challenging for instance segmentation. First, ADE20K contains plenty of small objects. It is hard for most of instance segmentation frameworks to distinguish small objects from background, and even harder to recognize and classify them into correct categories. Second, ADE20K is highly diverse in terms of scenes and objects, requiring models of strong capability to achieve better performance in various scenes. Third, scenes in ADE20K are generally crowded. The inter-class occlusion and intro-class occlusion create problems for object detection as well as instance segmentation. This is can be seen in fig.~\ref{qual_instance}, where the models struggle to detect some of the boxes in the cluttered areas (row 2, left) or the counter inf row 4, covered by multiple people.  

%To provide insights on how to address instance segmentation on images of such nature, we describe the main techniques used by the top submission, Megvii (Face++), as described by one of the authors.
To further gain insight from the insiders, we invite the leading author of the winner for the instance segmentation track in Places Challenge to give a summary of their winning method as follows: 

Following a top-down instance segmentation framework, ~\cite{peng2018megdet} starts with a module to generate object proposals first then classify each pixel within the proposal. But unlike RoI Align used in Mask-RCNN~\cite{he2017mask}, they use Precise RoI Pooling~\cite{jiang2018acquisition} to extract features for each proposal. Precise RoI Pooling avoids sampling the pivot points used in RoI Align by regarding a discrete feature map as a continuous interpolated feature map and directly computing a two-order integral. The good alignment of features provide with good improvement for object detection, while even higher gain for instance segmentation. To improve the recognition of small objects, they make use of contextual information by combining, for each proposal, the features of the previous and following layers. Given that top-down instance segmentation relies heavily on object detection, the model ensembles multiple object bounding-boxes before fed into a mask generator. We also find that the models cannot avoid predicting objects in the mirror, which indicates that current models are still incapable of high-level reasoning in parallel with low-level visual cues.

\section{Object-Part Joint Segmentation}
\label{section:jointtraining}

\begin{figure*}
\begin{center}
\includegraphics[width=1.\textwidth, trim={0 0 0 0}, clip]{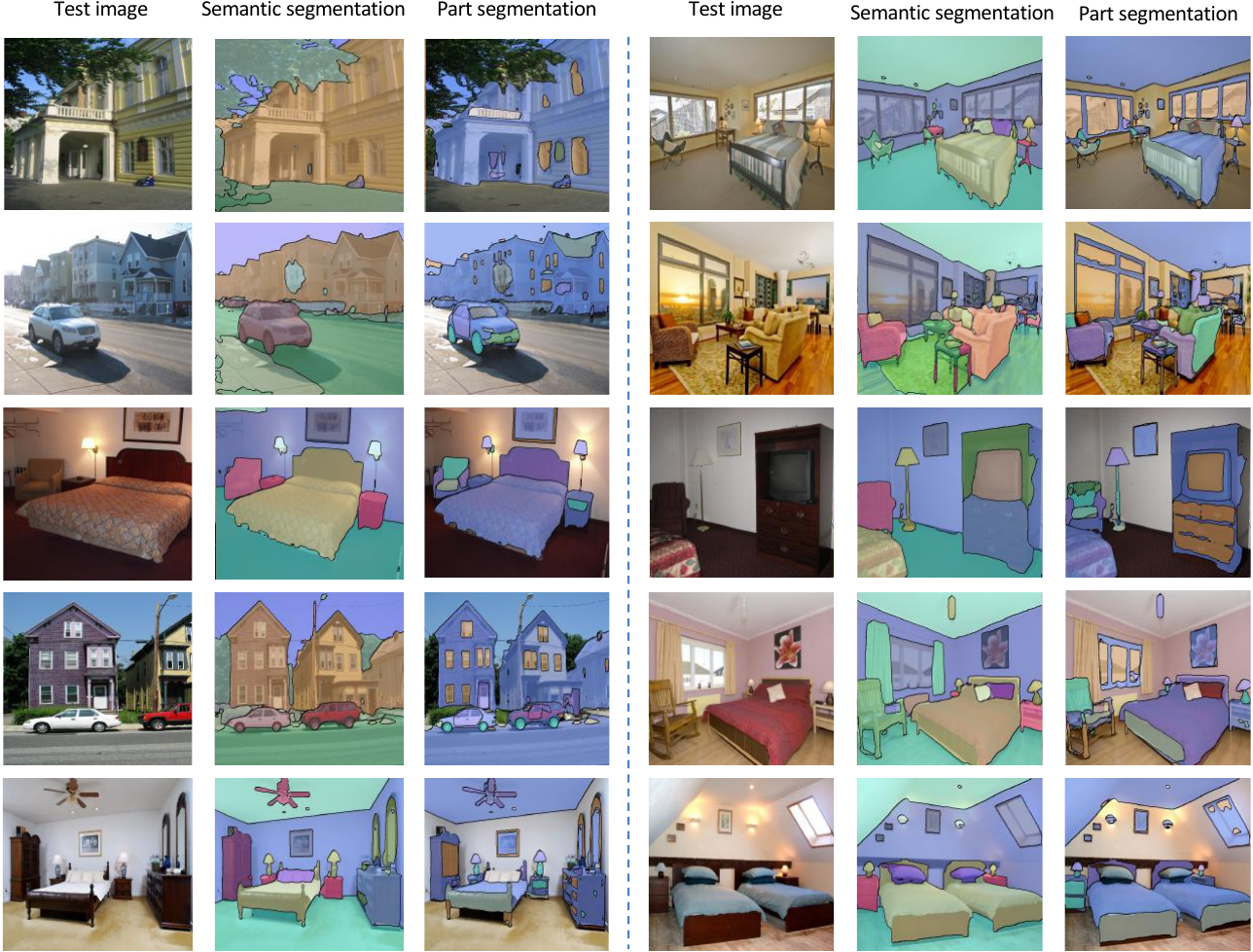}
\end{center}
\caption{Object and part joint segmentation results predicted by UPerNet. Object parts are segmented based on the top of the corresponding object segmentation mask.}
\label{part_result}
\end{figure*}

\begin{figure}
\begin{center}
\includegraphics[width=0.45\textwidth, trim={0 0 0 0}, clip]{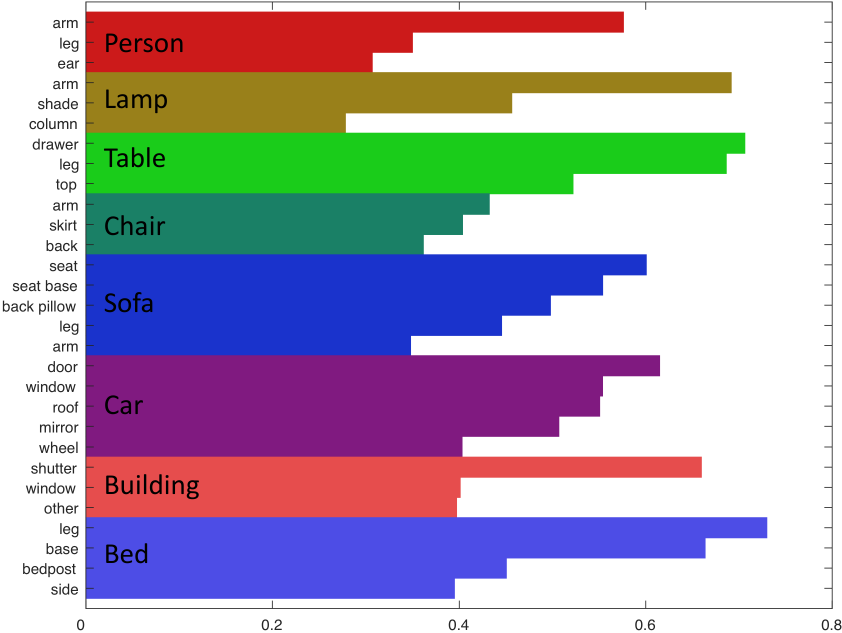}
\end{center}
\caption{Part segmentation performance (in pixel accuracy) grouped by several selected objects predicted by UPerNet.}
\label{part_acc}
\end{figure}

Since ADE20K contains part annotations for various object classes, we further train a network to jointly segment objects and parts. There are 59 out of total 150 objects that contain parts, some examples can be found in Fig.~\ref{fig:tree_fig}. In total there are 153 part classes included. We use UPerNet~\cite{xiao2018unified} to jointly train object and part segmentation. During the training, we include the non-part class and only calculate softmax loss within the set of part classes via ground-truth object class. During the inference, we first pick out a predicted object class, then get the predicted part classes from its corresponding part set. This is organized in a cascaded way. We show the qualitative results of UPerNet in Fig.~\ref{part_result}, and the quantitative performance of part segmentation for several selected objects in Fig.~\ref{part_acc}.

\iffalse

Since ADE20K also contain the part annotation for several object classes, we can further train network to segment parts on top of object segmentation. We select 8 objects with parts annotated (\textit{person}, \textit{lamp}, \textit{table}, \textit{chair}, \textit{sofa}, \textit{car}, \textit{building}, \textit{bed}), in total 36 part classes included in the training set. We found that training part segmentation directly by treating \textit{non-part} pixels as a \textit{background} class would bias the network towards \textit{background}. Cascade framework proposed in \cite{zhou2017scene} fits this task well by attending to the part-related objects first, and then performing part segmentation, more details can be found in~\cite{zhou2017scene}. We show Cascade-DilatedVGG model's qualitative results in Fig.~\ref{part_result}, and its quantitative performance for several selected classes in Fig.~\ref{part_acc}.

\textcolor{red}{ADD Object-Part Joint Training, Fig.16 and Fig.17 need redoing by adding the comparison of separate training and joint training}
\fi

\section{Applications}
\label{section:applications}

\begin{figure}
\begin{center}
\includegraphics[width=0.48\textwidth]{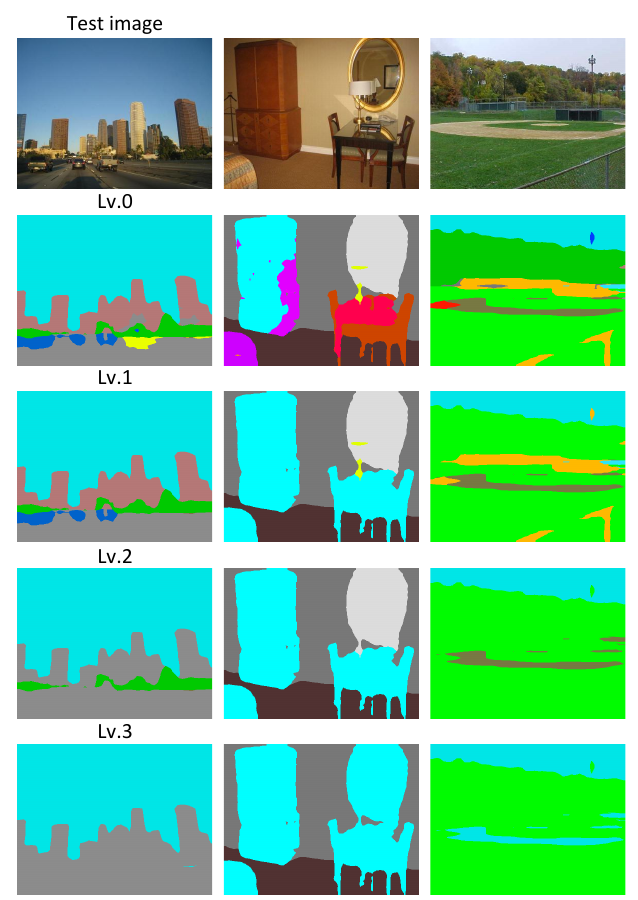}
\end{center}
\vspace{-4mm}
\caption{The examples of the hierarchical semantic segmentation. Objects with similar semantics like furnitures and vegetations are merged at early levels following the wordnet tree.}
\label{The hierarchical_segmentation_example}
\vspace{-3mm}
\end{figure}
 
\begin{figure}
\includegraphics[width=0.48\textwidth]{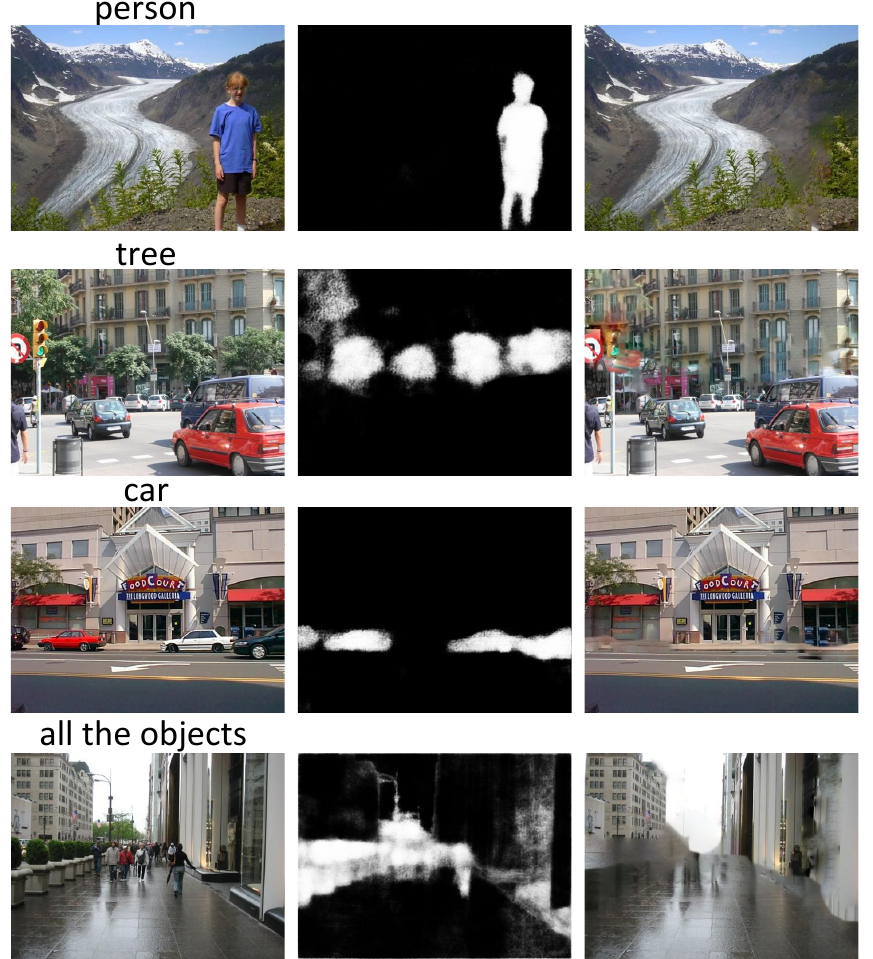}
\caption{\small Automatic image content removal using the predicted object score maps given by the scene parsing network. We are not only able to remove individual objects such as person, tree, car, but also groups of them or even all the discrete objects. For each row, the first image is the original image, the second is the object score map, and the third one is the filled-in image.}
\vspace{-3mm}
\label{image_completion}
\end{figure}

\begin{figure}
\begin{center}
\includegraphics[width=0.5\textwidth]{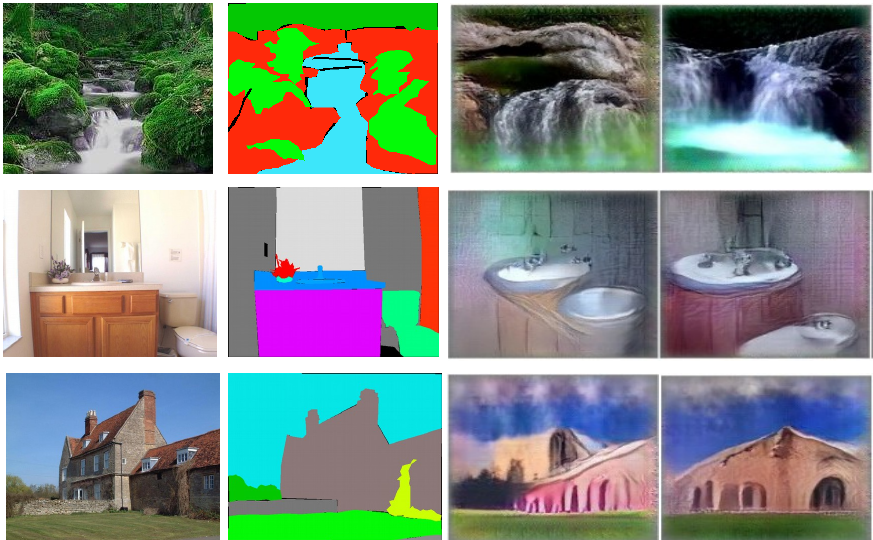}
\end{center}
\caption{Scene synthesis. Given annotation masks, images are synthesized by coupling the scene parsing network and the image synthesis method proposed in \cite{nguyen2016synthesizing}.}
\label{scene_synthesis}
\end{figure}

Accurate scene parsing leads to wider applications. Here we take the hierarchical semantic segmentation and the automatic scene content removal as exemplar applications of the scene parsing networks. 

\textbf{Hierarchical semantic segmentation}. Given the wordnet tree constructed on the object annotations shown in Fig.\ref{wordnet_tree}, the 150 categories are hierarchically connected and have hyponyms relations. Thus we could gradually merge the objects into their hyponyms so that classes with similar semantics are merged at the early levels. Through this way, we generated a hierarchical semantic segmentation of the image shown in Fig.~\ref{The hierarchical_segmentation_example}. The tree also provides a principled way to segment more general visual concepts. For example, to detect all furniture in a scene, we can simply merge the hyponyms associated with that synset, such as the \textit{chair}, \textit{table}, \textit{bench}, and \textit{bookcase}.

\textbf{Automatic image content removal}. Image content removal methods typically require the users to annotate the precise boundary of the target objects to be removed. Here, based on the predicted object probability map from scene parsing networks, we automatically identify the image regions of the target objects. After cropping out the target objects using the predicted object score maps, we simply use image completion/inpainting methods to fill the holes in the image. Fig.~\ref{image_completion} shows some examples of the automatic image content removal. It can be seen that with the object score maps, we are able to crop out the objects from an image precisely. The image completion technique used is described in ~\citep{huang2014image}.

\textbf{Scene synthesis}. Given a scene image, the scene parsing network could predict a semantic label mask. Furthermore, by coupling the scene parsing network with the recent image synthesis technique proposed in \cite{nguyen2016synthesizing}, we could also synthesize a scene image given the semantic label mask. The general idea is to optimize the input code of a deep image generator network to produce an image that highly activates the pixel-wise output of the scene parsing network. Fig.~\ref{scene_synthesis} shows three synthesized image samples given the semantic label mask in each row. As comparison, we also show the original image associated with the semantic label mask. Conditioned on an semantic mask, the deep image generator network is able to synthesize an image with similar spatial configuration of visual concepts.

\section{Conclusion}
\label{section:conclusion}
In this work we introduced the ADE20K dataset, a densely annotated dataset with the instances of stuff, objects, and parts, covering a diverse set of visual concepts in scenes. The dataset was carefully annotated by a single annotator to ensure precise object boundaries within the image and the consistency of object naming across the images. Benchmarks for scene parsing and instance segmentation are constructed based on the ADE20K dataset. We further organized challenges and evaluated the state-of-the-art models on our benchmarks. All the data and pre-trained models are released to the public.

%A generic network design called Cascade Segmentation Module was proposed for scene parsing. It enables the convolutional neural networks to parse scenes into stuff, objects, and object parts in cascade with the state-of-the-art performance.

\begin{acknowledgements}
This work was supported by Samsung and NSF grant No.1524817 to AT. SF acknowledges the support from NSERC. BZ is supported by a Facebook Fellowship. 
\end{acknowledgements}

\bibliographystyle{spbasic}      % basic style, author-year citations
\bibliography{egbib}


\end{document}